\documentclass[lettersize,journal]{IEEEtran}
\usepackage{amsmath,amsfonts}
\usepackage{amsthm}
\usepackage{amssymb}
\usepackage{algorithmic}
\usepackage{algorithm}
\usepackage{array}
\usepackage{textcomp}
\usepackage{url}
\usepackage{verbatim}
\usepackage{graphicx}
\usepackage{cite}
\usepackage{booktabs}
\usepackage{xspace}
\usepackage{multirow}
\usepackage{hyperref}

\newcommand{\ie}{i.e.}

\newcommand{\model}{AGREE\xspace}
\DeclareMathOperator*{\argmin}{arg\,min}

\newtheorem{remark}{Remark}

\usepackage{subcaption}
\usepackage{xcolor}
\usepackage[switch]{lineno}
\definecolor{mycustom_red}{HTML}{CC1717}

\begin{document}

\author{Xinxi Chen,
        Junyang Chen,
        Yiqun~Zhang,~\IEEEmembership{Senior Member,~IEEE},
        Chuangming Qiu,
        and Xiang Zhang

\thanks{Xinxi Chen and Yiqun Zhang are with the School of Computer Science and Technology, Guangdong University of Technology, Guangzhou 510006, China (e-mail: 3222004975@mail2.gdut.edu.cn, yqzhang@gdut.edu.cn). Junyang Chen is with Tsinghua University, Shenzhen 518055, China (e-mail: chenjuny25@mails.tsinghua.edu.cn). Chuangming Qiu and Xiang Zhang are with BH-Energy Technology Co., Ltd., Shenzhen 518000, China, and also with Shenzhen MSU-BIT University, Shenzhen 518172, China. (e-mail: \{13510302922, roger.j\}@139.com). (Corresponding author: Yiqun Zhang.)}

\thanks{\copyright~2026 IEEE. Personal use of this material is permitted. Permission from IEEE must be obtained for all other uses, in any current or future media, including reprinting/republishing this material for advertising or promotional purposes, creating new collective works, for resale or redistribution to servers or lists, or reuse of any copyrighted component of this work in other works. This is the author-accepted version. The final published version will be available on IEEE Xplore.}

}

\title{Bridge the Gaps: Heterogeneous Attributed Graph Clustering via Quaternion Representation Learning}

\markboth{IEEE Transactions on Emerging Topics in Computational Intelligence, June 2026}%
{Chen \MakeLowercase{\textit{et al.}}: Bridge the Gaps: Heterogeneous Attributed Graph Clustering via Quaternion Representation Learning}

\maketitle

\begin{abstract}
Attributed graph clustering partitions nodes by jointly exploiting node attributes and graph topology. It remains challenging due to attribute heterogeneity and representation degradation during graph learning. 
Real-world datasets often contain heterogeneous attributes (\ie, numerical and categorical attributes), complicating unified representation learning. This challenge becomes more complex in attributed graphs, where constructing a clustering-friendly graph structure from attributes and topology remains difficult. Under deep graph architectures, repeated graph propagation causes node embeddings to become overly similar, leading to the over-smoothing (OS) effect. Meanwhile, graph representation learning amplifies topological influence, making discriminative attribute information harder to exploit for clustering, an effect we refer to as over-dominating (OD).
To bridge these gaps,
an end-to-end framework, \underline{A}ny-type attributed \underline{G}raph \underline{RE}presentation l\underline{E}arning (\model) is proposed. It unifies attributed graphs and any-type attributed data through multi-level alignment and similarity-based graph construction. Quaternion-based graph convolution strengthens attribute interaction to alleviate OD, while shallow graph architectures help relieve OS. The learned embeddings are jointly optimized for graph reconstruction and clustering, without requiring a predefined number of clusters during training.
Experiments on diverse benchmarks show that \model achieves strong overall performance in accuracy, robustness, and adaptability.
\end{abstract}


\section{Introduction}\label{sec:introduction}

\begin{figure}
    \centering
    \includegraphics[width=1\linewidth]{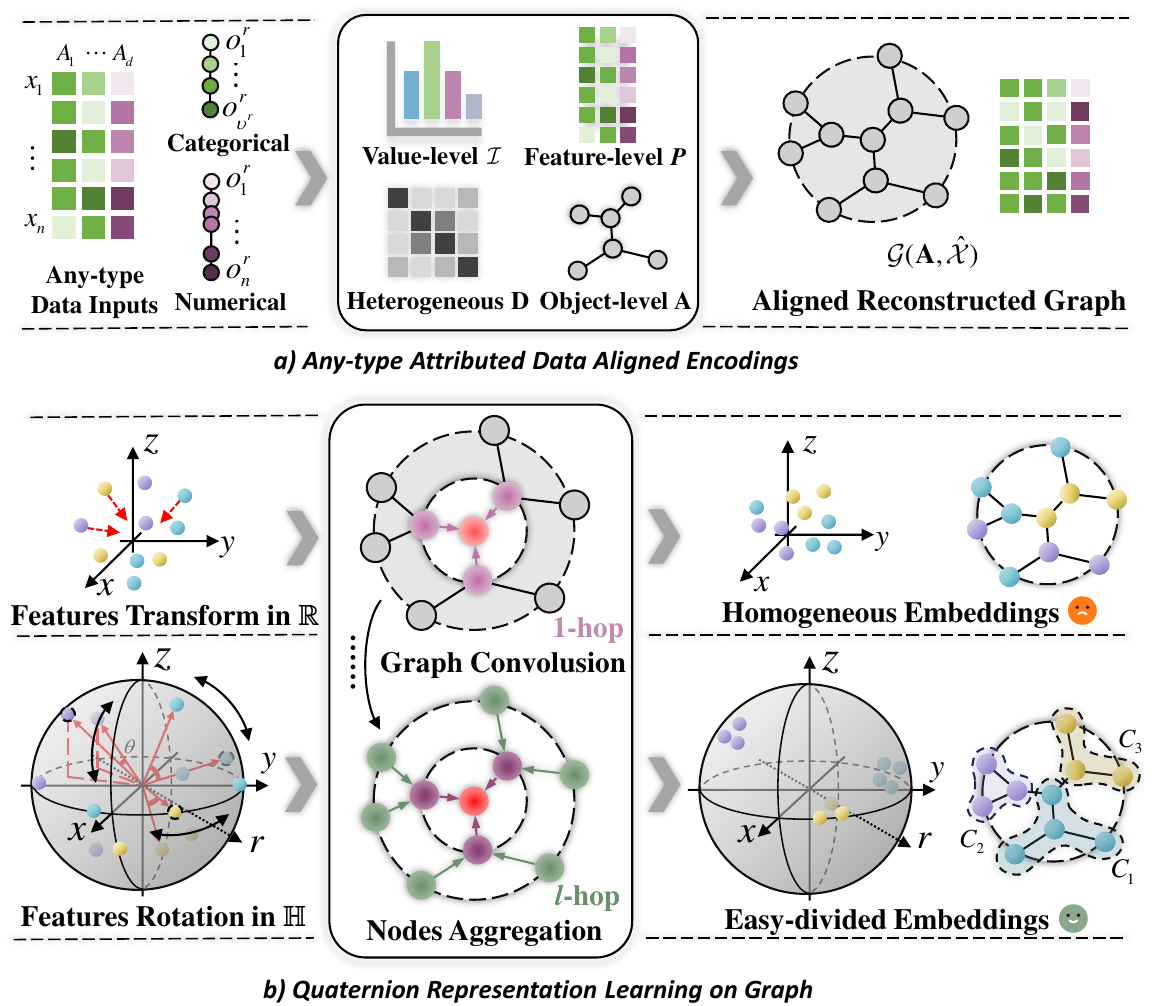}
    \caption{
    Motivation and design of \model. a) Multi-level aligned encoding and graph construction mitigate attribute heterogeneity by organizing heterogeneous attributes into a unified graph representation. b) In contrast to vanilla graph encoders, quaternion graph encoders strengthen cross-view interaction in hyper-complex space, helping alleviate the OS and OD effects and thus improving clustering quality.
    }
    \label{fig:intro}
\end{figure}

\IEEEPARstart{C}{lustering} is a key unsupervised machine learning technique in knowledge discovery and data mining, with the goal of uncovering hidden distributions of data objects.
Effective clustering enables the identification of latent device behaviors, anomaly patterns, and functional groupings, essential for optimizing resource allocation, enhancing security, and interpreting complex sensor-driven environments.  
In attributed graphs, data objects are treated as nodes and are equipped with descriptive attributes while edges capture relational information, allowing algorithms to integrate attribute and topological information for richer, more discriminative representations. 
Accordingly, attributed graph clustering approaches, especially deep graph neural networks, have been proposed to learn fused representations from attributes and graph topology \cite{EGAE, ARGAE/ARVGAE,CCGC}. Recent research has also started to consider more diverse data types, and studies clustering for complex data in a more general setting~\cite{cdc}.
However, heterogeneous attributes and graph neural networks still introduce a series of gaps that limit both the performance and application of current approaches.

From the model perspective, deep learning-based graph representation, particularly those based on Graph Convolution Networks (GCN) \cite{GCN}, have proven to be effective in boosting the accuracy of attributed graph clustering. These methods, along with their variants \cite{DAEGC, EGAE}, simultaneously capture the graph’s structural information and node attributes, resulting in more comprehensive and insightful embeddings. This is also consistent with studies suggesting that topology-centered propagation may be insufficient in heterogeneous graph learning~\cite{latgrl}. In practice, global relationship interaction is important when exploring the cluster distribution of graph data. Increasing the number of graph convolution layers can help capture long-range dependencies. However, it often leads to the \emph{Over-Smoothing} (OS) effect, where node representations become overly similar due to excessive feature overlap.

Most OS solutions focus more on relieving the smoothing of nodes in terms of topological information for serving supervised learning tasks. However, in unsupervised clustering, overemphasizing the graph topology and the distinction of nodes may hamper the learning of compact clusters. That is, the similarity of nodes' embeddings can be easily \emph{dominated} by the topology while attribute information tends to be overlooked as the graph convolution operation is guided by the topology. As a result, such an \emph{Over-Dominating} (OD) effect will improperly drive the learning process away from the objective of clustering. 
Intuitively, a wide and shallow network can emphasize the coupling of attributes without incurring OS. However, ``shallow'' restricts abstract-level representations, and ``wide'' limits model scalability and efficiency, resulting in a contradiction when simultaneously solving OS and OD.
These challenges are further exacerbated in dense graphs reconstructed from attribute similarities, whose adjacency matrices contain similarity values rather than binary edges. This increased density causes \emph{faster} propagation of information, but also makes the graph more susceptible to OS and OD effects. The emphasis on graph topology can overshadow the importance of attribute information, leading to less distinct node embeddings. Consequently, this can hinder the learning of compact clusters and reduce clustering performance.
Therefore, alleviating OS and OD is essential for learning clustering-friendly representations on attributed graphs.

From the perspective of real-world data complexity, analyzing heterogeneous attributed data comprising categorical and numerical attributes is challenging due to the attribute type gap. That is, numerical features describe the data quantitatively, whereas categorical features provide qualitative descriptions, making unified similarity measurement difficult. 
Recent efforts have therefore sought to better handle this attribute-type heterogeneity through richer similarity design and more adaptive clustering strategies, including value occurrence frequency, semantic ordinal relationships, and inter-feature dependencies~\cite{Zhang2025ComplexIntelligentSystems}.
Compared with traditional approaches (e.g., one-hot encoding and Hamming distance \cite{Hamming}), which only match values directly, advanced metrics define distance structures based on statistical distributions of feature values. Recent works \cite{ZhangC23agraph,h2h,tnn/ZhangCT20,pami/ZhuCY22} further exploit statistical priors on feature coupling to better represent categorical data.
However, they mainly address attribute-type heterogeneity,
while neglecting the construction of an appropriate graph representation that captures inter-instance relationships from heterogeneous attributes and graph topology.

To bridge the heterogeneous-attributes-level and attribute-graph-level gaps, 
graph convolution is performed in a four-axis hyper-complex space to strengthen attribute interaction and thereby alleviate OD.
It adopts a shallow model architecture, which also counters OS. More specifically, the hyper-complex space $\mathbb{H}$ is introduced for representation learning because the Hamilton product enables
efficient feature rotation and interaction~\cite{parcollet2020survey}. 
As illustrated in Fig.~\ref{fig:intro}, the feature quaternion can be transformed by the parameter quaternion to model feature interaction.
Compared with real-valued space $\mathbb{R}$, quaternion transformation provides four times the Degrees of Freedom (DoF) under the same parameter scale, thereby strengthening coupling learning without introducing excessive parameters.
This stronger representation capacity enhances the contribution of attributes to representation learning and helps alleviate the OD effect. At the same time, the strong fitting capability of quaternion operation supports expressive shallow graph learning, which further reduces the need for deep GCN stacking and thus lowers the risk of OS.

As for the overall model design, a graph auto-encoder framework is adopted to perform \textbf{A}ny-type Attributed \textbf{G}raph \textbf{Re}presentation L\textbf{e}arning (\textbf{\model}). It uses aligned encoding to construct the graph representation of mixed-type attributed graph data through value-level, feature-level, attribute-type level, and object-level coupling encoding, and then processes it using hyper-complex representation learning. We map input attributes into the hyper-complex space $\mathbb{H}$, utilizing four views that correspond to the four components of a quaternion, enabling efficient feature quaternion transformation via the model's learnable parameter quaternion. This four-view feature amplification also strengthens the role of attributes, helping to mitigate the over-dominance (OD) effect. By training the parameters of the Four-View Projection (FVP) and Quaternion Graph Encoders (QGE) using a loss function that integrates graph reconstruction loss and a clustering objective, \model produces embeddings optimized for a variety of clustering tasks. Since an ideal $k$ is typically not available in real-world IoT clustering scenarios, a general clustering objective is integrated into the loss function without requiring a predefined $k$. This enables the model to learn more versatile representations and, crucially, ensures its practicality in most scenarios where the 'true' $k$ is unknown. Theoretical analysis and extensive experiments on various real benchmark graph datasets have illustrated the superiority of \model. The source code is available at \url{https://github.com/XinxiiChen/AGREE}. The main contributions are summarized as:
\begin{itemize}
    \item \textbf{Generalized attributed graph clustering framework.} 
    \model is proposed as a unified representation learning framework for both attributed graphs and mixed-type attributed data. It aligns
    heterogeneous numerical and categorical attributes and constructs a suitable graph representation for subsequent learning, and also extends quaternion algebra to the modeling of arbitrary-dimensional attributes in an unsupervised learning setting. 
    \item \textbf{Identification and formulation of the OD effect.} 
    The \emph{Over-Dominating (OD)} effect in graph clustering is identified and analyzed, revealing how excessive emphasis on topology can lead to underutilization of attribute information. 
    This provides a novel perspective for explaining performance bottlenecks and 
    for analyzing clustering cases with relatively less reliable topology.
    \item \textbf{Quaternion-based graph representation for OS and OD mitigation.} 
    A quaternion-based graph representation learning scheme is introduced to strengthen structured attribute interaction on aligned heterogeneous attributes. It alleviates OD by enhancing attribute-side modeling during graph propagation, while its shallow yet expressive architecture helps reduce the risk of OS.
    \item \textbf{Flexible representation learning across cluster granularities.}
    \model learns versatile embeddings without being tied to a predefined $k$ in advance. These embeddings support multiple cluster granularities and reduce the need for retraining when exploring different cluster resolutions.
\end{itemize}

\section{Related works}\label{sec:related-works}
\subsection{Mixed-type Attributed Data Clustering}
Existing methods for clustering mixed-type attributed data fall into two categories: 1) designing distance measures tailored to categorical features, and 2) transforming data into numerical representations for clustering.

Recent studies on categorical data clustering have improved distance calculations by exploiting statistical properties of the data. Value similarity can be modeled from a probabilistic perspective, where two values are considered more similar if their joint occurrence probability is higher. An information-theoretic view \cite{tnn/ZhangCT20} instead uses entropy derived from probability distributions to measure dissimilarity. Recent work further improves distance learning by learning cluster-aware value relationships directly from data \cite{Zhao2026BreakTheTie} and by exploiting order information in distance construction \cite{AAAI/Zhang}. Moreover, work \cite{ZhangC23agraph} unifies the treatment of numerical, nominal, and ordinal features, improving the universality of clustering.
Another line of research transforms categorical values into numerical forms while preserving information useful for clustering. 
Recent studies develop learnable distance construction strategies. Work \cite{pami/zhang_intra_attribute_distance} learns intra-attribute semantics for distance construction, while \cite{Zhao2024ECAI} exploits order information in qualitative attributes. More recently, the work \cite{Zhang2025PACMMOD} learns value relationships in a clustering-oriented manner, making distance construction more adaptive to clustering tasks. Recent work also improves robustness through multi-granular competitive learning \cite{Cai2024ICDCS}.

For heterogeneous data with different attribute types, effective clustering further requires compatible representations and distance measures. Recent work \cite{Zhang2025ESWA} addresses this issue through unified metric learning. An alternative perspective is to exploit the complementary information provided by different attribute types or encoded representations. Accordingly, SSA-SNMF~\cite{mohammadi2025adaptive} adaptively learns similarity matrices for multi-view clustering, while FMvPCI~\cite{yang2025fmvpci} combines multiple representations with fuzzy clustering. On this basis, some multi-view clustering methods further pursue unified representation learning across views \cite{Zhang2024DeepURL}, whereas others explicitly learn graph structures to support clustering \cite{Chen2024SparseGraphTensor}. External semantic priors, including those from large language models, may also provide a promising direction for heterogeneous attribute modeling.
However, these strategies often rely on prior assumptions about the data or representation design, which may limit their adaptability. The resulting representations can then be directly used by conventional clustering algorithms such as K-Means-type methods or spectral clustering.

\subsection{Attributed Graph Clustering}
Attributed graph clustering aims to partition nodes with both topological links and attribute information into compact groups, and has attracted increasing attention. Existing methods can be roughly grouped into three categories: 1) graph autoencoder-based methods, 2) contrastive learning-based methods, and 3) explicit modeling-based methods.

Subsequent studies further extend this reconstruction-based paradigm with attention mechanisms \cite{DAEGC}, adversarial regularization \cite{ARGAE/ARVGAE}, or joint optimization objectives, as in EGAE \cite{EGAE}. DFCN \cite{DFCN} integrates autoencoder and graph autoencoder features via a fusion module, and employs triplet self-supervision for robust clustering. HyReaL~\cite{chen2026hyreal} further extends this line by exploring richer hyper-complex representation spaces. Generative or refinement-oriented designs, including diffusion-based methods, have also been explored to improve graph representation quality.
These methods are effective in representation reconstruction and benefit from relatively stable training and clustering regularization.

To further improve representation discrimination, contrastive learning has become another important paradigm, usually accompanied by additional view construction or auxiliary contrastive objectives. CCGC \cite{CCGC} leverages clustering as a proxy task to enhance representation discrimination. CONVERT \cite{mm/convert} introduces a reversible perturb–recover mechanism to preserve semantic consistency across augmented views. More recently, GLAC \cite{GLAC-TAI} jointly exploits local and global topologies via dual GCN pipelines with adaptive contrastive learning; SCGDN \cite{SCGDN-mm} proposes an augmentation-free framework using attentional diffusion; and MAGI \cite{MAGI-kdd} bridges contrastive learning and modularity maximization by defining community-aware contrastive signals. These methods are well suitable for noisy or structurally challenging graph settings.

{Other recent studies form another line of research by adopting more explicit optimization or modeling formulations. Some of them emphasize general clustering optimization or structure learning. CDC~\cite{cdc} provides a general clustering
framework for complex data with an anchor-based design, while GE-S-D~\cite{Tang2024AnchorDebias} further incorporates debiasing to reduce structural bias in attributed graph clustering. By contrast, DESE~\cite{DeSE} emphasizes structure
learning guided by entropy minimization. Another branch follows probabilistic or matrix-factorization formulations. In this line, LCAAG~\cite{yang2025link} adopts a Bayesian formulation, whereas ERNMF~\cite{mohammadi2025dual} and
RENMF~\cite{berahmand2025relative} follow matrix-factorization formulations. These methods usually offer clearer modeling assumptions and stronger interpretability. However, they are often less flexible for learning complex graph-attribute
interactions in an end-to-end manner.}
\subsection{Quaternion Representation Learning}
Quaternions are a four-dimensional extension of complex numbers, grounded in a comprehensive algebraic structure. The Hamilton product \cite{quaternion-Hamilton-product} effectively manages the interaction among the four quaternion elements through vector rotation along the three imaginary axes, positioning quaternion operations as a promising tool for improving representation learning capabilities \cite{parcollet2020survey}. This is especially relevant for data with naturally interconnected feature components. The study \cite{nisp2019-quaternion-KG} is recognized as the pioneering effort to apply quaternions to knowledge graph embedding.

Currently, most quaternion applications occur within supervised learning frameworks. The input data often feature inherent tuples or multiple interdependent components that correspond to the three imaginary parts of the quaternion, making them particularly suitable for representation learning via quaternion rotation operations. However, their use in more general unsupervised clustering settings remains limited.

\section{Preliminary}\label{sec:preliminary}

This section introduces the definition of mixed-type attributes graph data. 
A mixed-type attributed graph is a data structure that extends traditional graphs by associating each node with attributes of various types, including numerical and categorical features. 

For an undirected attributed graph $G=\{\mathbf{A,X}\}$, the $\mathbf{A}\in\mathbb{R}^{n\times n}$ is the adjacency matrix representing the relationships between nodes and $\mathbf{X}\in\mathbb{R}^{n\times d}$ is the node attribute matrix. 
A degree matrix $\mathbf{D}\in\mathbb{R}^{n\times n}$ reflects the connectivity of each node, where $\mathbf{D}_{ii} = \sum^{n}_{j=1} \mathbf{A}_{ij}$ is the sum of edges connected to node $i$. To normalize the adjacency matrix, we compute the symmetric normalized Laplacian matrix $\mathbf{\tilde{A}=D^{-\frac{1}{2}} \tilde{L} D^{-\frac{1}{2}}}$ where $\mathbf{\tilde{L}}=\mathbf{I}+\mathbf{A}$ is the self-loop adjacency matrix, and $\mathbf{I}$ is the identity matrix.
The clustering goal is to partition the nodes into $k$ clusters $\{G_1, G_2, ..., G_k\}$ based on both the structural and attribute information.

Besides, a mixed-type attribute dataset $\mathcal{H}$ that can be reconstructed as a graph structure is formally represented as an attributed triplet set:
\begin{equation}
    \mathcal{H}=\mathcal{<X,A,O>},
\end{equation}
where $\mathcal{X}=\{\boldsymbol{X}_l|l=1,2,\dots,n\}$ represents the set of $n$ samples, and each sample $\boldsymbol{X}_l=[x_l^1,x_l^2,\dots,x_l^d]^\top$. Also, $\mathcal{A}=\{\boldsymbol{A}^r|r=1,2,\dots,d\}$ is the set of $d$ feature values, where each feature $\boldsymbol{A}^r=[a_1^r,a_2^r,\dots,a_n^r]^\top$ is associated with $n$ node-specific values. Finally, $\mathcal{O}=\{\mathcal{O}^1,\mathcal{O}^2,..., \mathcal{O}^d\}$ represents the set of possible value sets, where $\mathcal{O}^r=\{o^r_1,o^r_2,\dots,o^r_{\upsilon^r}\}$ for categorical features. 

In this work, the feature set $\mathcal{A}$ is divided into two subsets, \ie, categorical and numerical features. Specifically, let $d^{\{c\}}$ and $d^{\{u\}}$
denote the numbers of categorical and numerical features, respectively. Then, $\mathcal{A}=\mathcal{A}^{\{c\}}\cup\mathcal{A}^{\{u\}}$ and
$d=d^{\{c\}}+d^{\{u\}}$, where $\mathcal{A}^{\{c\}}$ and $\mathcal{A}^{\{u\}}$ denote the sets of categorical and numerical features, respectively. For a
categorical feature $\boldsymbol{A}^r\in\mathcal{A}^{\{c\}}$, its values are drawn from a finite set $\mathcal{O}^r$, and $\upsilon^r$ denotes the number of
distinct values in $\boldsymbol{A}^r$. By contrast, each feature in $\mathcal{A}^{\{u\}}$ takes continuous real-valued observations. Based on these two types of
features, a more informative graph structure is constructed for mixed-type data clustering.

The quaternion operations used in the hyper-complex domain are introduced below. A quaternion $Q$ is a hyper-complex number in the field $\mathbb{H}$ and can be written as:
\begin{equation}
 Q=r+x\boldsymbol{\rm{i}}+y\boldsymbol{\rm j}+z\boldsymbol{\rm k}, 
\end{equation}
where $r$ is for the real part and $x\boldsymbol{\rm{i}}+y\boldsymbol{j}+z\boldsymbol{\rm k}$ represents the imaginary parts. In $\mathbb H$, an orthogonal relationship holds for the imaginary parts, \ie, $\boldsymbol{\rm {i^{2}}}=\boldsymbol{\rm{j^{2}}}=\boldsymbol{\rm {k^{2}}}=\boldsymbol{\rm {ijk}}=-1$. 
The interaction between two quaternions $Q_1$ and $Q_2$ is defined by the Hamilton product, which is given by
\begin{equation}
    \begin{aligned}
        Q_1\otimes Q_2 
        &=(r_1r_2-x_1x_2-y_1y_2-z_1z_2) \\ 
        &+(r_1x_2 + x_1r_2 + y_1z_2-z_1y_2)\boldsymbol{\rm i} \\
        &+(r_1y_2-x_1z_2 + y_1r_2 + z_1x_2)\boldsymbol{\rm j} \\
        &+(r_1z_2 + x_1y_2-y_1x_2 + z_1r_2)\boldsymbol{\rm k}. 
    \end{aligned}
\end{equation}
This operation enables efficient interactions among feature components in the quaternion field, thereby providing greater flexibility for representing complex data encodings.

\section{Method}\label{sec:method}

\begin{figure*}
    \centering
    \includegraphics[width=0.85\linewidth]{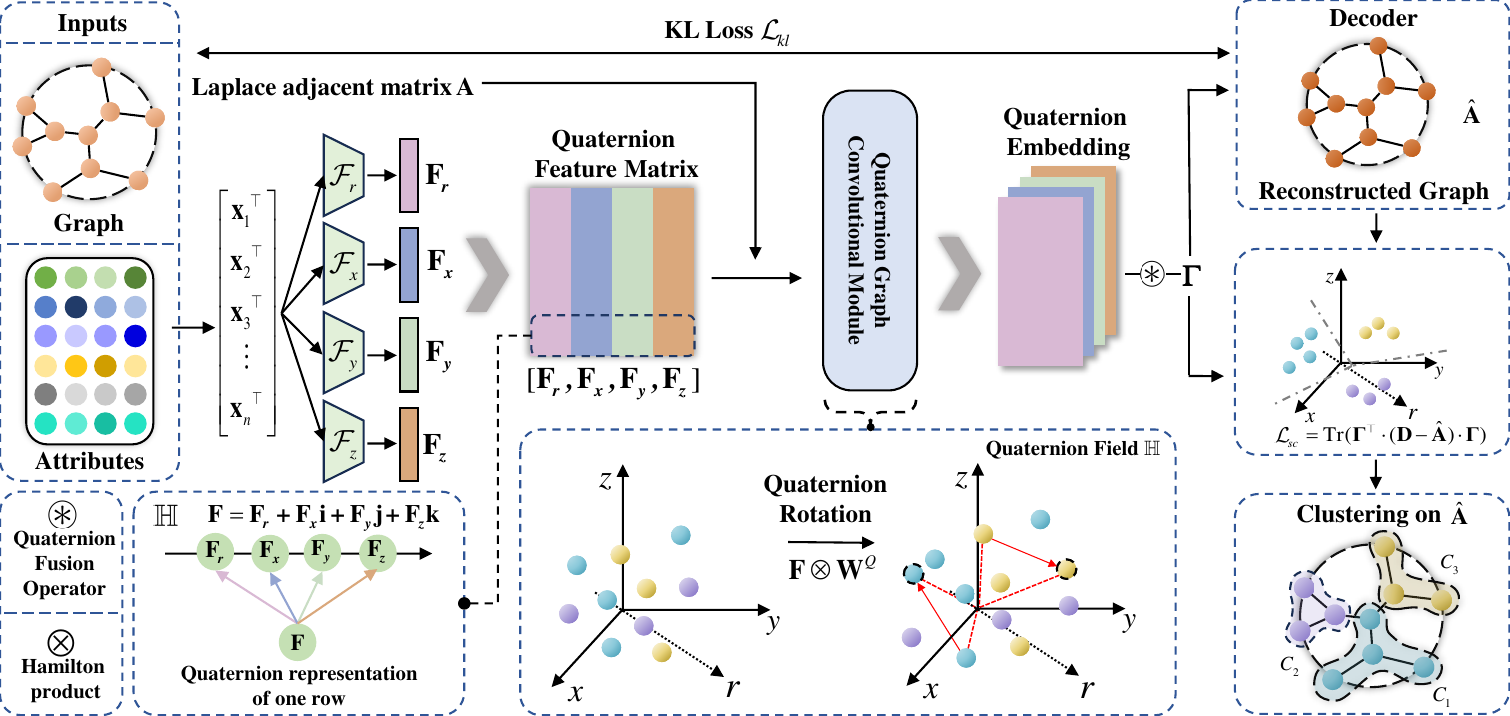}
    \caption{Overview of \model. {Mixed-type attributes are aligned and organized into an attributed graph $G=\{\mathbf{A},\hat{\mathcal{X}}\}$, in which the aligned attributes $\hat{\mathcal{X}}$ are projected into four views to form a feature quaternion $\mathbf{F}_{\mathrm{prj}} = \mathbf{F}_\mathrm{r} + \mathbf{F}_\mathrm{x} \mathbf{i} + \mathbf{F}_\mathrm{y} \mathbf{j} + \mathbf{F}_\mathrm{z} \mathbf{k}$. This representation is then encoded with the graph structure by quaternion graph convolution, through which the final embedding $\mathbf{U}$ is learned under joint graph reconstruction and clustering objectives for clustering.}}
    \label{fig:net}
\end{figure*}

This section first presents the Any-type Attributed Graph Data Aligned Encoding for aligning heterogeneous attributes with graph structure. It then introduces the Four-View Projection for organizing the aligned attributes into quaternion-compatible views, followed by the Hyper-Complex Graph Representation Learning framework for mixed-type attributed graph clustering. The overall pipeline is illustrated in Fig.~\ref{fig:net}.

\subsection{Any-type Attributed Graph Data Aligned Encoding}\label{subsec:AE}
Mixed-type attributed data contain numerical and categorical attributes. This makes direct graph-based representation learning difficult. Meanwhile, relationships among instances are important in many data analysis tasks. Such relationships can be naturally represented and exploited through graph structures and neighborhood aggregation. However, in many practical scenarios, the data are provided only as heterogeneous feature values rather than as an explicit graph. Therefore, these heterogeneous attributes need to be aligned into a unified representation and further organized into a graph format.
\subsubsection{Feature Aligned Encoding}
{The alignment in this subsection is driven by heterogeneity among attribute types. This differs from conventional multi-view alignment over predefined views and from heterogeneous graph learning with multiple node or edge types. Here, the main goal is to improve the comparability of mixed numerical and categorical attributes and to support graph construction for subsequent learning.}
There are four types of encoding in mixed-type attributed data: 1) value-level encoding, \ie, the relationships among the possible values within a categorical feature; 2) feature-level encoding, \ie, the dependencies between interrelated features; 3) attribute-type encoding, \ie, the interactions between different types of attributes in categorical features; and 4) object-level encoding, \ie, the relationships among data objects based on their similarities. By effectively encoding the data, the heterogeneous nature of mixed-type attributes is mitigated, and different attribute types are aligned through these encodings, enabling a more efficient learning process within a deep representation framework. In this subsection, we present the proposed encoding strategies for these encodings.

\paragraph{value-level encoding}
The probabilities associated with the potential values $\mathcal{O}^r = \{o^r_1, o^r_2, \dots, o^r_{\upsilon^r}\}$ that a categorical feature $\boldsymbol{A}^r$ can take on, can be regarded as a sequence of probabilities:
\begin{equation}\label{eq:value_coupling}
\mathcal{I}^r = \{P^{r}_i|i = 1, \dots, \upsilon^r\}.
\end{equation}
In this context, $P^{r}_i$ represents the occurrence probability of the possible value $o^r_i$ emerging within $\boldsymbol{A}^r$, and it is computed as follows:
\begin{equation}
    P^{r}_i = \frac{\delta(\{\boldsymbol{A}^r\}^n_1 = o^r_i)}{\delta(\{\boldsymbol{A}^r\}^n_1 \neq \text{Null})}.
    \label{intra-feature_p}
\end{equation}
Here, $\delta(\{\boldsymbol{A}^r\}^n_1 = o^r_i)$ tallies up the frequency at which $o^r_i$ appears in the feature value collection $\{\boldsymbol{A}^r\}^n_1$, while $\delta(\{\boldsymbol{A}^r\}^n_1 \neq \text{Null})$ enumerates the quantity of non-empty values within $\{\boldsymbol{A}^r\}^n_1$, which typically equates to $n$. It should be emphasized that the uppercase $P^{r}_i$ is employed here to demarcate the occurrence probability of a value from the conditional probability that will be introduced subsequently. Given that the probabilities adhere to the condition: 
  \[
  \sum_{i = 1}^{\upsilon^r} P^{r}_i = 1,
  \]
encoding the feature values based on the occurrence probabilities of their corresponding possible values undeniably enables the capture of value-level interconnections within each individual feature. 

\paragraph{feature-level encoding}\label{par:fea-encoding}
In real-world data sets, the original features generally exhibit a certain degree of interdependence. In order to depict such relationships between different features, we introduce the Conditional Probability Distribution (CPD) of a feature $\boldsymbol{A}^m$ conditioned on a possible value $o_i^r$ from another feature $\boldsymbol{A}^r$. This CPD is defined as a $\upsilon^m$-dimensional vector:
\begin{equation}
\boldsymbol{P}^{m|r}_i = [p(o_1^m|o_i^r), p(o_2^m|o_i^r), \dots, p(o_{\upsilon^m}^m|o_i^r)]^{\top},
\label{eq:P_vector}
\end{equation}
where the conditional probability $p(o_{j}^m|o_i^r)$ is calculated as:
\begin{equation}
p(o_{j}^m|o_i^r) = \frac{\sigma(\mathcal{X}^m_j \cap \mathcal{X}^r_i)}{\sigma(\mathcal{X}^r_i)}.
\end{equation}
Here, $\mathcal{X}^r_i = \{\boldsymbol{x}_l|x^r_l = o^r_i, l = 1, 2, \cdots, n\}$ denotes a subset of $\mathcal{X}$ consisting of all data objects whose $r$-th values are equal to $o^r_i$. The function $\sigma(\cdot)$ denotes the cardinality of a set. Through $\boldsymbol{P}^{m|r}_i$, the value $o^r_i$ can be encoded with respect to different features $\boldsymbol{A}^m \in \mathcal{A}^{\{c\}}$, thereby preserving the dependence among different features. 

\paragraph{attribute-type encoding}\label{par:attribute-encoding}
The feature-level encoding described above processes categorical features uniformly based on the Conditional Probability Distributions (CPDs). However, for heterogeneous feature data, numerical and categorical features still follow different distance structures. To bridge this gap while preserving their inherent properties, categorical feature values are projected onto a set of one-dimensional spaces. The categorical values are then encoded according to their projected positions.

\begin{remark}\label{rmk:unification}
    \textbf{Connecting mixed-type Features.} The motivation behind projecting categorical values onto one-dimensional spaces is to unify categorical and numerical features. By doing so, both types of features can represent distances in a consistent way, thus establishing the foundation for appropriately aligning mixed-type features.
\end{remark} 

The projection is carried out using a commonly-employed categorical feature distance metric based on the CPDs defined in Section~\ref{par:fea-encoding}. For a feature $\boldsymbol{A}^r$, the distance between two possible values $o^r_i$ and $o^r_j$ is calculated with respect to each of the categorical features $\mathcal{A}^{\{c\}}$ and can be expressed as:
\begin{equation}
    d(o^r_i,o^r_j)=\sum\limits_{\boldsymbol{A}^m\in\mathcal{A}^{\{c\}}}  \left\lVert\boldsymbol{P}_i^{m|r}-\boldsymbol{P}_j^{m|r}\right\rVert.
\end{equation}

With this distance definition, all $\upsilon^r$ possible values of feature $\boldsymbol{A}^r$ can be projected onto each of the $\upsilon^r(\upsilon^r-1)/2$ one-dimensional spaces spanned by corresponding pairs of possible values. Specifically, for a one-dimensional space $\mathcal{R}^r_{ij}$ spanned by two possible values $o^r_i$ and $o^r_j$, the projection position of a value $o^r_t$ can be determined from its distance to $o^r_i$ (or $o^r_j$) in $\mathcal{R}^r_{ij}$ according to:
\begin{equation}
\phi(o^r_t,o^r_i;\mathcal{R}^r_{ij})=\frac{|d(o^r_t,o^r_i)^2-d(o^r_t,o^r_j)^2 + d(o^r_i,o^r_j)^2|}{2\cdot d(o^r_i,o^r_j)}.
\end{equation} 
This projection is derived from the Pythagorean theorem, and more details can be found in \cite{h2h}. After all $\upsilon^r$ possible values are projected, the
  pairwise distances between them in $\mathcal{R}^r_{ij}$ can be obtained. These distances are then organized into a symmetric matrix $\mathbf{D}^r_{ij}\in
  \mathbb{R}^{\upsilon^r \times \upsilon^r}$, where the $(t,l)$-th entry $\mathbf{D}^r_{ij}(t,l)$ denotes the distance between $o^r_t$ and $o^r_l$ in the projection
  space $\mathcal{R}^r_{ij}$.

\begin{remark}
    \textbf{Projection Comprehensiveness.} Each categorical feature $\boldsymbol{A}^r$ is represented by $\upsilon^r(\upsilon^r-1)/2$ one-dimensional structures, each corresponding to a different pair of possible values and thus providing a distinct endogenous view of the feature. As indicated in Remark~\ref{rmk:unification}, this projection preserves the internal relationships among possible values. Meanwhile, the one-dimensional representation makes it easier to establish a homogeneous connection with numerical features.
\end{remark}

\subsubsection{Coupling Encoding}\label{sct:encode_coupling}
Through the three types of encoding described above, all categorical features $\mathcal{A}^{\{c\}}$ are mapped to higher-dimensional spaces. More precisely, for the $l$-th value of a categorical feature $\boldsymbol{A}^r$ where $a^r_l = o^r_i$, $a^r_l$ will be encoded into a vector. This is achieved by concatenating its three types of encoding in the following way:
\begin{equation}\label{eq:three_encoding}
    \hat{\boldsymbol{a}}_l^r = [ P^r_i,\underbrace{\boldsymbol{P}^{1|r}_i,\boldsymbol{P}^{2|r}_i, \dots}_{d^{\{c\}}}, \underbrace{\mathbf{D}^r_{11}(i,\cdot),\mathbf{D}^r_{12}(i,\cdot),\dots}_{\upsilon^r(\upsilon^r-1)/2}].
\end{equation}
Here, $P^r_i$ represents the value-level encoding, the sequence $\boldsymbol{P}^{1|r}_i,\boldsymbol{P}^{2|r}_i,\dots$ corresponds to the feature-level encoding, and $\mathbf{D}^r_{11}(i,\cdot),\mathbf{D}^r_{12}(i,\cdot),\dots$ denotes the attribute-type encoding. The notation $\mathbf{D}^r_{ij}(t,\cdot)$ refers to the $t$-th row of the matrix $\mathbf{D}^r_{ij}$ defined in Section~\ref{par:attribute-encoding}.

By encoding each feature value within $\mathcal{A}^{\{c\}}$, the encoded set of categorical features can be denoted as $\mathcal{\hat{A}}^{\{c\}}$. Consequently, the entire feature set is updated to $\mathcal{\hat{A}}=\mathcal{\hat{A}}^{\{c\}}\cup\mathcal{A}^{\{u\}}$. Correspondingly, the object set associated with $\mathcal{\hat{A}}$ is denoted as $\mathcal{\hat{X}}$.
So far, three types of encodings have been introduced, namely value-level, feature-level, and attribute-type encodings. 

The final type, object-level encoding, is accomplished by constructing a fully-connected graph among the data objects. This will be discussed separately in the subsequent subsection. 

\subsubsection{Graph Aligned Construction}\label{subsubsec:graph-construction}
To construct the aligned graph representation, the object-level distance between two objects $\boldsymbol{x}_a$ and $\boldsymbol{x}_b$ is defined based on the $\ell_2$ norm as follows:
\begin{equation}\label{eq:object_dist}
\Psi(\boldsymbol{x}_a,\boldsymbol{x}_b)=\left\lVert[\Phi^{1}(x_{a}^{1},x_{b}^{1}),\Phi^{2}(x_{a}^{2},x_{b}^{2}),\dots,\Phi^{d}(x_{a}^{d},x_{b}^{d})]^\top\right\rVert_2.
\end{equation}
In this equation, $\Phi^{r}(x_{a}^{r},x_{b}^{r})$ denotes the distance induced by the $r$-th feature. To obtain a more reasonable distance measure for mixed-type data, the graph-based unified dissimilarity introduced in \cite{ZhangC23agraph} is adopted to define $\Phi^{r}(x_{a}^{r},x_{b}^{r})$.

For a categorical feature $\boldsymbol{A}^r\in \mathcal{A}^{\{c\}}$, let $x_{a}^{r}=o^r_i$ and $x_{b}^{r}=o^r_j$. Then, the distance $\Phi^{r}(x_{a}^{r},x_{b}^{r})$ is defined as:
\begin{equation}
    \Phi^{r}(x_{a}^{r},x_{b}^{r}) = 
    \begin{cases}
        \begin{aligned}
            & \sum_{m = 1}^{d} \phi^{r|m}(o^r_i,o^r_j)\cdot\omega^{r|m}, & \text{if } \boldsymbol{A}^r\in \mathcal{A}^{\{c\}} \\
            & |x^r_a-x^r_b|, & \text{if } \boldsymbol{A}^r \in \mathcal{A}^{\{u\}}
        \end{aligned}
    \end{cases}.
\label{eq:dissimilarity}
\end{equation}
Here, 
\begin{equation}
\phi^{r|m}(o^r_i,o^r_j)=\frac{\| |\boldsymbol{P}^{m|r}_i-\boldsymbol{P}^{m|r}_j| \|_1}{2}
\label{eq:F_function}
\end{equation}
denotes the dissimilarity between $o^r_i$ and $o^r_j$ as reflected by $\boldsymbol{A}^m$. The weight $\omega^{r|m}$ reflects the importance of $\boldsymbol{A}^m$ in determining the distances between values of $\boldsymbol{A}^r$. It can be specified by users or computed from the interdependence between $\boldsymbol{A}^m$ and $\boldsymbol{A}^r$.

{In line with \cite{ZhangC23agraph}, each numerical feature is discretized into five equal-width intervals, which provides a simple and moderately fine-grained partition without introducing excessive discretization complexity.}
The resulting interval indices are then treated as categorical values when computing Eq.~(\ref{eq:F_function}), since $\boldsymbol{A}^m$ may also be numerical. Although the value-level distances in the two cases of Eq.~(\ref{eq:dissimilarity}) take different forms, they can be unified from the perspective of transformation cost measured by the Earth Mover's Distance (EMD). Due to space limitations, readers are referred to \cite{ZhangC23agraph} for more details on the unification procedure and weight computation.

\begin{remark}\label{remark:Euclidean distance}
    \textbf{Rationale for Graph Aligned Construction.} Section \ref{sct:encode_coupling} yields an informative coupling-encoded set $\mathcal{\hat{X}}$, which can be used to compute object-level distances based on the Euclidean distance. However, graph construction is still performed on the original $\mathcal{X}$ using the graph-based unified dissimilarity for two main reasons: 1) it handles heterogeneous numerical and categorical features in a unified manner, thereby avoiding information loss; 2) the categorical features in $\mathcal{\hat{X}}$ are substantially expanded, so directly computing distances from $\mathcal{\hat{X}}$ would over-emphasize the categorical features.
\end{remark}

After computing the distances between all pairs of objects using Eq.~(\ref{eq:object_dist}), an adjacency matrix $\mathbf{A}\in\mathbb{R}^{n \times n}$ is obtained, where the $(i,j)$-th entry is given by $\Psi(\boldsymbol{x}_i,\boldsymbol{x}_j)$. Together with the coupling-encoded feature set $\mathcal{\hat{X}}$, the information derived from all four types of coupling relationships can be concisely represented by the constructed graph $\mathcal{G}=\{\mathbf{A}, \mathcal{\hat{X}}\}$. This graph enriches the internal information of the data and aligns it with the graph format, thereby laying the foundation for the subsequent mixed-type attributed graph clustering process.

\subsection{Hyper-Complex Graph Representation Learning}\label{subsec:graph-repre-learning}
{
To further model interactions among the aligned attributes, quaternion representation learning is introduced. The Hamilton product provides such a structured transformation in hyper-complex space. Unlike settings with naturally tuple-structured inputs or relational triplets, AGREE applies quaternion learning after mixed-type attribute alignment. Accordingly, the aligned attributes are projected into four views for subsequent quaternion graph encoding.
}

\subsubsection{Four-View Projection}
In contrast to many existing hyper-complex representation learning methods, which typically operate on tuple-structured feature components (e.g., RGB images), attributed graph data involves diverse attributes and varying graph structures. Therefore, a learnable projection mechanism is introduced to transform the aligned attribute matrix $\mathbf{X}$ into four distinct views. Here, $\mathbf{X}=\mathcal{\hat{X}}$, meaning that the attribute matrix used in this section is exactly the aligned encoding set obtained in the previous section. This mechanism removes the tuple-structure requirement on quaternion inputs and enables efficient coupling learning among attributes through the Hamilton product.

Specifically, the node attributes $\mathcal{\hat{X}}=\{\boldsymbol{\hat{X}}_l \mid l=1,2,\dots,n\}$ are projected by four independent MLPs into four views, \ie, $\mathbf{F}_\mathrm{r}$, $\mathbf{F}_\mathrm{x}$, $\mathbf{F}_\mathrm{y}$, and $\mathbf{F}_\mathrm{z}$. This projection is expressed as:
\begin{equation}
\mathbf{F}_{\mathrm{prj}} = \mathcal{F}_{\mathrm{prj}}(\mathcal{\hat{X}}) = \mathbf{W}^{\mathrm{L}}_{\mathrm{prj}}\mathcal{\hat{X}} + \mathbf{B}^{\mathrm{L}}_{\mathrm{prj}}, \quad {\mathrm{prj}} \in \{r, x, y, z\},
\label{eq:prj}
\end{equation}
where $\mathcal{F}_{\mathrm{prj}}(\cdot)$ denotes the MLP operation, and $\mathbf{W}^{\mathrm{L}}_{\mathrm{prj}}$ and $\mathbf{B}^{\mathrm{L}}_{\mathrm{prj}}$ represent the learnable parameters of the MLPs. These four views are combined to form a quaternion feature matrix $\mathbf{F} \in \mathbb{H}^{n \times (4 \times \hat{d})}$, expressed as:
\begin{equation}
\mathbf{F} = \mathbf{F}_\mathrm{r} + \mathbf{F}_\mathrm{x} \mathbf{i} + \mathbf{F}_\mathrm{y} \mathbf{j} + \mathbf{F}_\mathrm{z} \mathbf{k},
\label{eq:prj_concat}
\end{equation}
where $\hat{d}$ is the feature dimensionality encoded by $\mathcal{F}_\mathrm{prj}(\cdot)$. Each row of $\mathbf{F}$ corresponds to the quaternion representation of the respective node. Additionally, we introduce a learnable quaternion weight matrix $\mathbf{W}^{\mathrm{Q}} \in \mathbb{H}^{4d \times 4\hat{d}}$, with the same size as $\mathbf{F}$, to perform further transformation on $\mathbf{F}$:
\begin{small}
\begin{equation}
\mathbf{F} \otimes \mathbf{W}^\mathrm{Q} =
\left[
\begin{matrix}
\mathbf{F}_\mathrm{r}\\
\mathbf{F}_\mathrm{x}\\
\mathbf{F}_\mathrm{y}\\
\mathbf{F}_\mathrm{z}
\end{matrix}
\right ]^{\top}
\left[
 \begin{matrix}  
\mathbf{W}^\mathrm{Q}_\mathrm{r} & \mathbf{W}^\mathrm{Q}_\mathrm{x} & \mathbf{W}^\mathrm{Q}_\mathrm{y} &\mathbf{W}^\mathrm{Q}_\mathrm{z} \\
-\mathbf{W}^\mathrm{Q}_\mathrm{x} & \mathbf{W}^\mathrm{Q}_\mathrm{r} & -\mathbf{W}^\mathrm{Q}_\mathrm{z} & \mathbf{W}^\mathrm{Q}_\mathrm{y} \\
-\mathbf{W}^\mathrm{Q}_\mathrm{y} & \mathbf{W}^\mathrm{Q}_\mathrm{z} & \mathbf{W}^\mathrm{Q}_\mathrm{r} & -\mathbf{W}^\mathrm{Q}_\mathrm{x} \\
-\mathbf{W}^\mathrm{Q}_\mathrm{z} & -\mathbf{W}^\mathrm{Q}_\mathrm{y} & \mathbf{W}^\mathrm{Q}_\mathrm{x} & \mathbf{W}^\mathrm{Q}_\mathrm{r}
\end{matrix}
\right],
\label{eq:quaternion_transformation}
\end{equation}
\end{small}
where $\otimes$ represents the Hamilton product.

The processes described by $\mathcal{F}_\mathrm{prj}(\cdot)$ and $\otimes$ enable the transformation of any-dimensional attribute set into the quaternion space, while associating the feature components through the shared weights in the quaternion weight matrix $\mathbf{W}^\mathrm{Q}$. By adjusting the weights in $\mathbf{W}^\mathrm{Q}$, the features in $\mathbf{F}$ can be transformed with higher degrees of freedom (DoF), allowing the model to capture complex feature couplings more efficiently.

\begin{remark}
\textbf{Quaternion transformation enhances the Degree of Freedom (DoF).}
As shown in Eq.~(\ref{eq:quaternion_transformation}), the learnable parameters in $\mathbf{W}^\mathrm{Q}$, \ie, $\{\mathbf{W}^\mathrm{Q}_\mathrm{r}, \mathbf{W}^\mathrm{Q}_\mathrm{x}, \mathbf{W}^\mathrm{Q}_\mathrm{y}, \mathbf{W}^\mathrm{Q}_\mathrm{z}\}$, enable 16 different feature transformation pairs, facilitating the rotation of $\mathbf{F}$ within the hyper-complex space $\mathbb{H}$. In contrast, achieving the same transformation in real-value space $\mathbb{R}$ would require four times as many parameters. Thus, the DoF in quaternion feature transformation is four times greater than in $\mathbb{R}$. 
\label{rmk:dof}
\end{remark}

{
\begin{remark}
\textbf{Structured coupling and shallow expressive propagation in QGE for alleviating OD and OS.}
Compared with standard real-valued graph propagation $\tilde{A}XW$, QGE propagates quaternion features via $\tilde{A}H \otimes W^Q$. In real-valued networks, repeated multiplication by $\tilde{A}$ tends to homogenize node representations as depth increases~\cite{Oono2020}. In contrast, the Hamilton product $\otimes$ in
QGE introduces structured coupling among quaternion components, a property widely used in quaternion representation learning to model richer inter-component interactions~\cite{Tay2019}. This mechanism strengthens attribute-side modeling during propagation, making the learned representation less dependent on topology-
driven averaging and helping alleviate OD. Furthermore, the higher Degree of Freedom (DoF) of quaternion transformation provides richer expressive capacity. This allows AGREE to maintain strong representation learning with shallow propagation, reducing the need for deep chains that tend to homogenize embeddings, thereby
helping alleviate OS.
\label{rmk:qge-mechanism}
\end{remark}
}

\subsubsection{QGE: Quaternion Graph Encoders}
To integrate the feature quaternion $\mathbf{F}$ with the graph structure $\tilde{\mathbf{A}}$, the feature quaternion is passed through a quaternion graph convolution module composed of several stacked encoders. The operation of the $l$-th encoder can be expressed as:

\begin{equation}
\mathbf{H}_l = \varphi_l(\mathbf{\tilde{A}} \cdot \mathbf{H}_{l-1} \otimes \mathbf{W}^{Q}_l).
\label{eq:encoder}
\end{equation}

Here, $\otimes$ denotes the Hamilton product, which has higher precedence over the matrix multiplication operation. The function $\varphi_l(\cdot)$ represents the activation function, and $\mathbf{\tilde{A}} \in \mathbb{R}^{n \times n}$ is the normalized Laplacian adjacency matrix. Each encoder aggregates the $l$-hop quaternion representations of nodes based on the graph topology $\mathbf{\tilde{A}}$ to generate a more abstract feature representation $\mathbf{H}_l$. 

The final quaternion representation $\mathbf{H}_m$ is then fused into a single embedding matrix $\mathbf{U}$, calculated as:

\begin{equation}
\mathbf{U} = {\rm Re}(\mathbf{H}_m) \circledast {\rm Im}(\mathbf{H}_m),
\label{eq:avg}
\end{equation}
where ${\rm Re}(\cdot)$ and ${\rm Im}(\cdot)$ extract the real and imaginary parts of $\mathbf{H}_m$, respectively, and the operation $\circledast$ represents quaternion fusion, which averages the four parts of the quaternion feature. Finally, the graph is reconstructed as $\mathbf{\hat{A}} \in \mathbb{R}^{n \times n}$ using the embeddings $\mathbf{U}$ by:

\begin{equation}
\mathbf{\hat{A}} = \mathbf{U} \cdot \mathbf{U}^{\top}.
\label{eq:re}
\end{equation}

This reconstruction step acts as a decoder to preserve the original graph topology.

\subsection{Unified Optimization of Graph Reconstruction and Clustering}

The FVP and QGE modules work together to integrate attribute information within $\mathbf{U}$, with the graph reconstruction mechanism aligning this representation to the underlying graph structure, ensuring a balance between the attribute and graph consistency. To enhance clustering performance and promote sparsity in the reconstructed graph, a combined loss function is designed, incorporating the graph reconstruction loss $\mathcal{L}_{\mathrm{kl}}$, spectral clustering term $\mathcal{L}_{\mathrm{sc}}$, and a regularization term $\mathcal{L}_{\mathrm{reg}}$. The total loss is expressed as:
\begin{equation}
    \mathcal{L} = \mathcal{L}_{\mathrm{kl}} + \alpha\mathcal{L}_{\mathrm{reg}} + \beta\mathcal{L}_{\mathrm{sc}},
    \label{eq:loss}
\end{equation}
{where $\alpha$ controls the strength of the regularization term, and $\beta$ balances the contributions of the reconstruction-related term and the spectral clustering term.}
The reconstruction loss $\mathcal{L}_\mathrm{kl}$ measures the discrepancy between the original adjacency matrix $\mathbf{\tilde{A}}$ and the reconstructed adjacency $\mathbf{\hat{A}}$, and is defined as:
\begin{equation}
    \mathcal{L}_{\mathrm{kl}} = \frac{1}{n^2}\sum\limits_{i=1}^{n}\sum\limits_{j=1}^{n} \mathbf{\tilde{A}}_{ij}\log \frac{1}{\mathbf{\hat{A}}_{ij}},
\end{equation}
where $n$ is the number of nodes. Minimizing $\mathcal{L}_{\mathrm{kl}}$ enforces consistency between the graph topology encoded in $\mathbf{\tilde{A}}$ and the learned graph representation $\mathbf{\hat{A}}$. The regularization term $\mathcal{L}_{\mathrm{reg}}$ is introduced to prevent overfitting by limiting model complexity.

The spectral clustering loss $\mathcal{L}_{\mathrm{sc}}$ encourages $\mathbf{U}$ to align with the graph spectral structure and is defined as:
\begin{equation}
   \mathcal{L}_{\mathrm{sc}} = \mathrm{Tr} (\mathbf{U}^{\top} {\mathbf L} \mathbf{U}),
   \label{eq:loss-sc}
\end{equation}
where $\mathbf{L} = \mathbf{D} - \mathbf{\hat{A}}$ is the Laplacian matrix of the reconstructed graph, and $\mathbf{D}$ is the degree matrix of $\mathbf{\hat{A}}$.
According to the spectral clustering objective in Eq.~(\ref{eq:sc-process}), $\mathbf{U}$ in $\mathcal{L}_{\mathrm{sc}}$ can be viewed as a relaxed spectral embedding matrix. Minimizing $\mathcal{L}_{\mathrm{sc}}$ therefore encourages connected and highly similar nodes in $\mathbf{\hat{A}}$ to have similar embeddings, while favoring a graph structure with a lower cut cost, consistent with the spectral clustering objective.

After training, the clustering-friendly graph representation $\mathbf{\hat{A}}$ is obtained by Eq.~(\ref{eq:re}), and the corresponding semi-definite Laplacian
matrix $\mathbf{L} = \mathbf{D} - \mathbf{\hat{A}}$ is then constructed to support the spectral clustering objective:
\begin{equation}
\argmin_{\mathbf{H}}{\rm Tr}(\mathbf{H}^{\top} \mathbf{L} \mathbf{H})\ \ \ s.t.\ \ \mathbf{H}^{\top} \mathbf{H}=\mathbf{I}, 
\label{eq:sc-process}
\end{equation}
where $\mathbf{H} \in \mathbb{R}^{n \times k}$ represents the node-cluster affiliation matrix, and $\mathbf{I}$ is the identity matrix. 
Spectral clustering is performed by first calculating the eigenvectors of $\mathbf{L}$, then applying $K$-Means clustering to the eigenvector matrix. The number of clusters $k$ is a user-defined parameter.

\begin{table}[!t]
\centering
\caption{Statistics of the 19 datasets. $n$ and $k$ denote the number of nodes and clusters, while $d_u$ and $d_c$ represent the numerical and categorical feature dimensions, respectively.}
\label{tab:all-datasets}
\resizebox{\columnwidth}{!}{ 
\begin{tabular}{cllcccc}
\toprule
\textbf{No.} & \textbf{Dataset (Source)} & \textbf{Abbrev.} & \textbf{$n$} & \textbf{$d_u$} & \textbf{$d_c$} & \textbf{$k$} \\
\midrule
\multicolumn{7}{c}{\textit{\textbf{Attributed Graph Datasets}}} \\
\midrule
1  & ACM \cite{DBLP_ACM}               & ACM      & 3025 & 1870 & - & 3 \\
2  & Wiki-CS \cite{wiki}                  & WIKI     & 2405 & 4973 & - & 17 \\
3  & CiteSeer \cite{cora_citeseer}     & CITE & 3327 & 3703 & - & 6 \\
4  & DBLP \cite{DBLP_ACM}              & DBLP     & 4057 & 334  & - & 4 \\
5  & FILM \cite{GeomGCN} & FILM & 7600 & 932  & - & 5 \\
6  & Cora \cite{cora_citeseer}         & CORA     & 2708 & 1433 & - & 7 \\
7  & Wisconsin \cite{GeomGCN}  & WISC  & 251  & 1703 & - & 5 \\
8  & USA Air-Traffic \cite{githubdataset/deep_graph_clustering_survey}   & UAT   & 1190 & 239  & - & 4 \\
9 & Amazon Photo \cite{githubdataset/deep_graph_clustering_survey}  & AMAP  & 7650 & 745  & - & 8 \\
\midrule
\multicolumn{7}{c}{\textit{\textbf{Mixed-Type Attributed Datasets}}} \\
\midrule
10 & Mammographic \cite{UCI_Repository}      & MM  & 961   & 1  & 4  & 2 \\
11 & Heart Failure \cite{UCI_Repository}     & HF  & 299   & 7  & 5  & 2 \\
12 & Breast Cancer \cite{UCI_Repository}     & BC  & 286   & 4  & 5  & 2 \\
13 & Autism-Adolescent \cite{UCI_Repository} & AA  & 104   & 2  & 7  & 2 \\
14 & Tic-Tac-Toe \cite{UCI_Repository}       & TTT & 958   & -  & 9  & 2 \\
15 & Zoo \cite{UCI_Repository}               & ZO  & 101   & -  & 16 & 7 \\
16 & Yeast \cite{UCI_Repository}             & YE  & 1484  & 8  & -  & 10 \\
17 & Glass Identification \cite{UCI_Repository}    & GI  & 214   & 9  & -  & 6 \\
18 & Wine \cite{UCI_Repository}              & WI  & 178   & 13 & -  & 3 \\
19 & Iris \cite{UCI_Repository}              & II  & 150   & 4  & -  & 3 \\
\bottomrule
\end{tabular}
}
\end{table}

{
\begin{remark}
\textbf{Complexity Analysis.} \label{remark:complexity-analysis}
Let $n$ be the number of nodes, $d$ the number of original attributes, $\hat{d}$ the projected feature dimension, $m$ the number of QGE
layers, and $|E|$ the number of edges in the input graph. Let $\upsilon^r$ denote the number of possible values of the $r$-th categorical attribute, and let $V=\max_r \upsilon^r$. The overall time complexity of AGREE is $O\!\left(C_{\mathrm{align}} + n d \hat{d} + m|E|\hat{d} + n^2 \hat{d}\right)$, where $C_{\mathrm{align}}$ denotes the cost of the aligned encoding stage. The total complexity consists of three parts.
\\
\indent 1) \textit{Aligned encoding and graph construction:} The value-level encoding requires counting the occurrence frequency of each categorical value, which is linear in the data size. The feature-level encoding computes conditional probability distributions among attribute values. For the attribute-type encoding, the distance between each pair of categorical values is computed from the conditional probability vectors in Eq.~(\ref{eq:P_vector}). Let $V$ be the maximum categorical cardinality. Since there are $O(V^2)$ value pairs and each distance computation costs $O(V)$, the complexity of this step is $O(d^2V^3)$. In addition, object-level graph construction computes pairwise distances between nodes according to Eq.~(\ref{eq:object_dist}), which introduces an $O(n^2 d)$ term in the dense case. Therefore, the aligned encoding stage is dominated by
$C_{\mathrm{align}} = O(d^2 n + d^2 V^3 + n^2
d)$.
\\
\indent 2) \textit{Four-view projection and quaternion graph encoding:} The four-view projection maps $\mathcal{\hat{X}}$ into $\mathbf{F}_{\mathrm{r}}$, $\mathbf{F}_{\mathrm{x}}$, $\mathbf{F}_{\mathrm{y}}$, and $\mathbf{F}_{\mathrm{z}}$ by four MLPs, which costs $O(n d\hat{d})$. The QGE module then performs $m$ layers of graph propagation on quaternion features. For a graph with $|E|$ edges, the propagation cost is $O(m|E|\hat{d})$. The quaternion fusion in Eq.~(\ref{eq:avg}) is linear in the embedding size and is therefore dominated by the graph propagation stage.
\\
\indent 3) \textit{Graph reconstruction and clustering optimization:} The reconstructed adjacency matrix $\mathbf{\hat{A}}=\mathbf{U}\mathbf{U}^{\top}$ in Eq.~(\ref{eq:re}) requires $O(n^2 \hat{d})$ time in the dense case. The computation of $\mathcal{L}_{\mathrm{kl}}$ and the Laplacian-based spectral term in Eq.~(\ref{eq:loss-sc}) is of the same order as the adjacency-related operations. Hence, this stage is dominated by the dense reconstruction cost $O(n^2 \hat{d})$.
\\
\indent The overall memory complexity is dominated by storing adjacency-related matrices, which is $O(n^2)$ in the dense case.
\\
\indent It should be noted that $O(d^2V^3)$ is a worst-case upper bound for attribute-type encoding. In practice, categorical cardinalities are usually moderate, so this term does not dominate the actual overhead. Omitting it, the overall time complexity reduces to $O(d^2n + n^2d + nd\hat{d} + m|E|\hat{d} + n^2\hat{d})$. For larger-scale applications, sparse or blockwise graph operations can further reduce the practical time and memory costs.
\end{remark}
}

\section{Experiments}\label{sec:experiment}
This section presents the experimental evaluation of \model on two types of benchmark datasets, i.e., attributed graph datasets and mixed-type attributed datasets. A total of 19 datasets and 13 compared
methods are included to assess the effectiveness of \model.

\subsection{Experimental Settings}
\subsubsection{Datasets}

\begin{table*}[!t]
\centering
\caption{Clustering performance on the Mixed-Type Attributed Datasets evaluated by ACC, NMI, and ARI (Mean$\pm$Std). The best and second-best results on each dataset are marked in \textbf{bold} and \underline{underline}, respectively. `AR' denotes the average performance rank of each method.}
\resizebox{1\linewidth}{!}{
\begin{tabular}{c|c|rrrrrrrrrrrrrr}
\toprule
    \multirow{2}{*}{Dataset}   
  & \multirow{2}{*}{Metric}  
  & \multicolumn{1}{c}{$K$-Means} 
  & \multicolumn{1}{c}{GAE}  
  & \multicolumn{1}{c}{ARVGAE\cite{ARGAE/ARVGAE}}
  & \multicolumn{1}{c}{DAEGC\cite{DAEGC}} 
  & \multicolumn{1}{c}{CCGC\cite{CCGC}} 
  & \multicolumn{1}{c}{DFCN\cite{DFCN}} 
  & \multicolumn{1}{c}{EGAE\cite{EGAE}} 
  & \multicolumn{1}{c}{CONVERT\cite{mm/convert}} 
  & \multicolumn{1}{c}{SCDGN\cite{SCGDN-mm}} 
  & \multicolumn{1}{c}{MAGI\cite{MAGI-kdd}} 
  & \multicolumn{1}{c}{GLAC\cite{GLAC-TAI}}
  & \multicolumn{1}{c}{{DESE\cite{DeSE}}}
  & \multicolumn{1}{c}{{CDC\cite{cdc}}}
  & \multicolumn{1}{c}{\textbf{\model}} \\
  
  &
  & \multicolumn{1}{c}{(Classical)}
  & \multicolumn{1}{c}{(Classical)}
  & \multicolumn{1}{c}{(IJCAI'18)} 
  & \multicolumn{1}{c}{(IJCAI'19)} 
  & \multicolumn{1}{c}{(AAAI'23)} 
  & \multicolumn{1}{c}{(AAAI'21)} 
  & \multicolumn{1}{c}{(TNNLS'22)} 
  & \multicolumn{1}{c}{(MM'23)} 
  & \multicolumn{1}{c}{(MM'23)} 
  & \multicolumn{1}{c}{(KDD'24)} 
  & \multicolumn{1}{c}{(TAI'25)} 
  & \multicolumn{1}{c}{{(KDD'25)}}
  & \multicolumn{1}{c}{{(TNNLS'25)}}
  & \multicolumn{1}{c}{\textbf{(Ours)}} \\

\midrule
\multirow{3}{*}{MM}
        & ACC & 68.55±0.00 & 53.37±0.00 & 55.59±1.00 & 53.06±5.00 & 68.31±2.00 & 68.55±0.00 & 68.52±1.00 & 67.54±1.05 & 59.65±5.57 & 52.30±1.28 & 51.57±0.00 & {\underline{78.75±5.92}} &{ 58.84±0.13} & \textbf{82.96±0.00} \\
        & NMI & 11.02±0.00 & 0.76±0.00 & 1.03±1.00 & 1.05±3.00 & 10.09±1.00 & 10.65±0.00 & 10.30±1.00 & 10.57±0.34 & 4.88±5.26 & 0.40±0.53 & 0.25±0.00 & {\underline{27.25±9.67}} & {6.93±0.15} & \textbf{34.87±1.00} \\
        & ARI & 13.67±0.00 & 0.35±0.00 & 1.17±1.00 & 1.09±3.00 & 13.38±2.00 & 13.67±0.00 & 13.62±1.00 & 12.25±1.05 & 4.86±5.57 & 0.16±1.28 & 0.01±0.00 & {\underline{34.38±11.81}} & {8.74±0.19} & \textbf{43.40±1.00} \\
\midrule
\multirow{3}{*}{HF}
        & ACC & 62.21±0.00 & 63.61±1.00 & 68.90±0.00 & 67.89±1.00 & 63.55±0.00 & 57.42±2.00 & \underline{69.00±2.00} & 60.20±2.88 & 53.41±0.70 & 55.72±4.77 & 68.23±0.00 & {63.55±4.63} & {60.54±0.09} & \textbf{71.37±3.00} \\
        & NMI & 0.25±0.00 & 0.74±1.00 & 2.67±0.00 & 1.01±1.00 & 0.75±0.00 & 0.02±0.00 & \underline{2.96±3.00} & 0.24±0.31 & 0.04±0.01 & 0.95±1.64 & 1.17±0.00 & {0.37±0.24} & {1.70±0.01} & \textbf{10.92±3.00} \\
        & ARI & 1.75±0.00 & 3.36±0.00 & 3.31±0.00 & 0.85±1.00 & \underline{3.37±0.00} & -0.04±0.00 & 2.63±336.00 & 0.89±2.88 & -0.15±0.70 & 1.18±4.77 & 0.74±0.00 & {1.36±0.86} & {0.18±0.03} & \textbf{16.63±4.00} \\
\midrule
\multirow{3}{*}{BC}
        & ACC & 51.40±0.00 & 70.98±0.00 & 71.54±1.00 & 69.93±0.00 & 70.98±1.00 & 55.31±5.00 & 66.47±6.00 & 66.26±4.46 & 54.20±0.00 & \underline{72.03±0.49} & \textbf{72.24±1.63} & {70.42±3.62} & {65.45±0.11} & \underline{72.03±0.00} \\
        & NMI & 0.21±0.00 & 6.84±0.00 & \underline{7.56±1.00} & 0.39±0.00 & 7.53±1.00 & 1.66±2.00 & 2.66±3.00 & 4.63±2.56 & 4.11±0.00 & 4.30±1.26 & 6.30±2.76 & {5.93±2.94} & {5.92±0.03} & \textbf{8.66±1.00} \\
        & ARI & -0.25±0.00 & 14.68±0.00 & \underline{15.72±1.00} & -0.40±0.00 & 15.24±1.00 & 1.64±4.00 & 3.62±601.00 & 9.53±4.46 & -0.80±0.00 & 5.77±0.49 & 12.47±1.63 & {10.62±7.59} & {10.41±0.10} & \textbf{17.00±0.00} \\
\midrule
\multirow{3}{*}{AA}
        & ACC & 50.96±0.00 & 51.92±0.00 & 53.94±1.00 & 61.25±1.00 & 57.88±2.00 & 51.44±2.00 & 61.92±1.00 & 51.92±1.43 & 60.58±0.00 & 55.38±2.95 & \underline{62.50±1.05} & {59.42±2.46} & {61.54±0.03} & \textbf{63.56±1.00} \\
        & NMI & 0.06±0.00 & 0.14±0.00 & 0.02±0.00 & 2.50±0.00 & 1.54±0.00 & 0.06±0.00 & 2.71±1.00 & 0.32±0.43 & 0.85±0.00 & 0.72±1.03 & \underline{3.24±1.31} & {1.40±1.36} & {2.19±0.01} & \textbf{4.71±1.00} \\
        & ARI & -0.90±0.00 & -1.27±0.00 & -0.87±1.00 & 0.80±69.00 & 1.48±1.00 & -0.81±1.00 & 1.62±2.00 & -0.82±1.43 & 2.08±0.00 & 0.37±2.95 & \underline{2.49±1.05} & {0.54±1.96} & {2.47±0.03} & \textbf{3.75±2.00} \\
\midrule
\multirow{3}{*}{TTT}
        & ACC & 57.83±0.00 & 59.92±0.00 & 54.61±1.00 & 62.77±6.00 & 60.33±2.00 & 53.35±1.00 & 65.31±1.00 & 59.61±2.95 & 62.38±1.14 & 56.92±5.70 & 66.10±1.38 & {61.36±5.38} & {\underline{69.81±0.08}} & \textbf{95.74±3.00} \\
        & NMI & 0.74±0.00 & 1.10±0.00 & 0.11±1.00 & 0.14±0.00 & 1.26±1.00 & 0.25±0.00 & 0.84±1.00 & 2.09±1.17 & 1.35±0.09 & 1.79±2.56 & {2.57±2.80} & {0.79±0.79} & {\underline{11.25±0.09}} & \textbf{76.33±10.00} \\
        & ARI & 1.95±0.00 & 3.08±0.00 & 0.45±0.00 & -0.14±0.00 & 3.43±1.00 & 0.40±1.00 & 0.92±2.00 & 3.65±2.95 & 4.24±1.14 & 2.62±5.70 & 3.83±1.38 & {2.13±1.85} & {\underline{15.42±0.12}} & \textbf{83.81±9.00} \\
\midrule

\multirow{3}{*}{ZO}
        & ACC & 72.38±7.00 & 59.60±1.00 & 76.34±1.00 & 36.47±3.00 & 68.42±3.00 & 71.68±8.00 & 68.91±3.00 & 67.82±5.71 & \underline{77.92±0.63} & 40.10±2.35 & 53.96±8.83 & {72.67±0.79} & {77.03±0.16} & \textbf{80.20±2.00} \\
        & NMI & 77.61±5.00 & 54.79±2.00 & 73.15±3.00 & 15.34±7.00 & 63.40±2.00 & 70.68±5.00 & 71.11±3.00 & 66.45±2.29 & 73.99±1.14 & 26.00±3.27 & 41.57±11.60 & {81.88±2.39} & {\textbf{82.86±0.10}} & \underline{82.12±2.00} \\
        & ARI & 67.55±11.00 & 38.11±1.00 & 64.46±2.00 & -4.10±1.00 & 50.15±4.00 & 57.44±10.00 & 56.87±5.00 & 57.60±5.71 & \underline{75.13±0.63} & 10.76±2.35 & 26.73±8.83 & {66.24±1.80} & {73.39±0.22} & \textbf{77.33±3.00} \\
\midrule
\multirow{3}{*}{YE}
        & ACC & 36.20±1.00 & 23.23±1.00 & 28.49±0.00 & 31.27±0.00 & 22.40±3.00 & 21.74±3.00 & 36.38±1.00 & 27.84±1.54 & 31.02±0.77 & 19.86±2.36 & 31.32±0.16 & {\underline{38.52±1.95}} & {34.95±0.01} & \textbf{44.22±1.00} \\
        & NMI & \underline{26.42±1.00} & 8.74±0.00 & 8.95±0.00 & 1.10±0.00 & 7.58±2.00 & 16.14±1.00 & 25.75±1.00 & 10.81±3.51 & 21.63±0.25 & 2.53±0.62 & 1.15±0.13 & {17.42±1.70} & {\textbf{26.88±0.00}} & 21.82±1.00 \\
        & ARI & \underline{13.95±1.00} & 4.16±0.00 & 5.15±0.00 & 0.05±0.00 & 3.26±2.00 & 5.78±1.00 & \underline{13.95±1.00} & 2.99±1.54 & 10.34±0.77 & 0.15±2.36 & 0.05±0.16 & {11.46±2.76} & {13.87±0.00} & \textbf{14.31±1.00} \\
\midrule
\multirow{3}{*}{GI}
        & ACC & 54.21±0.00 & 43.46±0.00 & 39.02±1.00 & 36.17±3.00 & 44.77±1.00 & 45.56±1.00 & \underline{55.05±1.00} & 41.54±3.16 & 44.11±0.23 & 26.03±1.98 & 35.98±0.00 & {48.13±3.33} & {36.92±0.08} & \textbf{55.09±3.00} \\
        & NMI & \underline{42.07±1.00} & 30.16±0.00 & 19.49±1.00 & 6.15±6.00 & 30.03±2.00 & 31.48±2.00 & 37.63±1.00 & 23.30±4.15 & 29.63±0.10 & 4.38±0.78 & 3.22±0.00 & {35.48±1.86} & {17.76±0.07} & \textbf{50.04±6.00} \\
        & ARI & \underline{26.55±1.00} & 19.93±0.00 & 12.89±0.00 & 1.32±1.00 & 17.28±3.00 & 17.70±1.00 & 25.87±1.00 & 12.48±3.16 & 16.87±0.23 & -0.08±1.98 & -0.74±0.00 & {21.47±2.58} & {6.80±0.06} & \textbf{34.76±8.00} \\
\midrule

\multirow{3}{*}{WI}
        & ACC & 49.33±1.00 & 38.37±1.00 & 43.82±0.00 & 39.33±2.00 & 57.75±1.00 & 50.00±1.00 & 74.72±0.00 & 72.13±1.40 & 91.57±0.00 & 37.98±1.13 & 44.16±7.91 & {85.06±0.57} & {\underline{95.51±0.00}} & \textbf{96.46±0.00} \\
        & NMI & 11.23±1.00 & 2.79±1.00 & 5.00±0.00 & 2.21±1.00 & 16.29±3.00 & 8.75±1.00 & 36.55±1.00 & 40.31±1.85 & 72.41±0.00 & 0.77±0.38 & 9.26±14.80 & {57.68±1.31} & {\underline{85.29±0.00}} & \textbf{87.42±1.00} \\
        & ARI & 11.69±1.00 & -0.89±0.00 & 3.99±0.00 & -0.16±0.00 & 17.20±2.00 & 10.52±0.00 & 38.99±0.00 & 41.95±1.40 & 75.30±0.00 & -0.25±1.13 & 4.84±7.91 & {59.24±1.35} & {\underline{86.85±0.00}} & \textbf{89.45±1.00} \\
\midrule
\multirow{3}{*}{II}
        & ACC & 89.33±0.00 & 80.00±0.00 & 88.27±4.00 & 34.22±1.00 & 92.53±1.00 & 83.73±1.00 & 91.67±1.00 & 84.47±3.43 & \underline{94.67±0.00} & 38.80±2.53 & 56.80±12.28 & {67.47±5.89} & {70.00±0.13} & \textbf{96.27±2.00} \\
        & NMI & 75.82±0.00 & 56.84±0.00 & 76.54±5.00 & 2.67±1.00 & 77.56±2.00 & {70.12±1.00} & {77.86±2.00} & 64.13±4.37 & \underline{83.22±0.00} & 1.34±0.98 & 49.01±26.76 & {58.93±8.96} & {46.80±0.22} & \textbf{86.95±4.00} \\
        & ARI & 73.02±0.00 & 54.75±0.00 & 71.85±8.00 & 0.03±0.00 & 79.51±1.00 & {63.54±2.00} & {77.95±3.00} & 63.12±3.43 & \underline{85.08±0.00} & -0.02±2.53 & 36.22±12.28 & {49.06±4.71} & {40.36±0.23} & \textbf{89.23±4.00} \\
\midrule

        - & AR & {7.3} & {9.2} & {8.2} & {11.1} & {6.4} & {9.0} & {\underline{5.0}} & {8.1} & {6.8} & {11.3} & {8.8} & {5.6} & {6.1} & \textbf{{1.1}} \\
\bottomrule
\end{tabular}
}
\label{tab:mixed-data-compared-results}
\end{table*}

Experiments are conducted on two types of benchmark datasets, including nine \textit{\textbf{Attributed Graph Datasets}} and ten \textit{\textbf{Mixed-Type Attributed Datasets}}. The former consists of graph-structured data with node attributes, while the latter contains categorical and numerical attributes, including pure categorical and pure numerical cases. Statistics and sources of the datasets are shown in Table~\ref{tab:all-datasets}.

\subsubsection{Experiment Process}
All experiments are conducted using PyTorch 1.8.0 on an NVIDIA 3090 GPU. The model is first pre-trained for 10 epochs with only the KL loss $\mathcal{L}_{kl}$ and regularization loss $\mathcal{L}_{reg}$ for warm-up. Following recent graph clustering methods \cite{CCGC,EGAE,DAEGC,DFCN,mm/convert}, the clustering performance is reported as the average result with standard deviation over ten independent runs of each compared method. In each run, the model is trained for 50 epochs. 

\subsubsection{Compared Methods and \model Settings}
Thirteen clustering methods are included for comparison, including the classic $K$-Means and twelve state-of-the-art learning-based graph clustering approaches. These baselines are categorized into: 1) autoencoder-based methods, i.e., GAE, ARVGAE \cite{ARGAE/ARVGAE}, EGAE \cite{EGAE}, DAEGC \cite{DAEGC}, and DFCN \cite{DFCN}; 2) contrastive learning-based methods, i.e., CONVERT \cite{mm/convert}, CCGC \cite{CCGC}, SCDGN \cite{SCGDN-mm}, MAGI \cite{MAGI-kdd}, and GLAC \cite{GLAC-TAI}; and 3) {graph structure learning methods, i.e., CDC~\cite{cdc} and DESE~\cite{DeSE}.}
 
For most compared graph clustering methods on the mixed-type attributed datasets, categorical features are first transformed by one-hot encoding, and the adjacency matrix is then constructed based on the
inner product between the expanded features. 
For \model, the graph $\mathcal{G}=\{\mathbf{A},\hat{\mathcal{X}}\}$ produced by the proposed aligned encodings is used as input. The quaternion graph encoder is implemented with two quaternion graph convolution layers.

The implementation details and hyper-parameter settings are summarized as follows. For GAE, 200 epochs of unsupervised training are first performed, and $K$-Means is then applied to the learned embeddings. The remaining compared methods are implemented based on their source code. 
For \model, the main settings include the model architecture and training procedure. The layer dimensions are set to $[512, 256, 128]$. Adam is adopted as the optimizer, and ReLU is used as the activation function in the graph encoder. The pre-training learning rate is set to $10^{-4}$, and the model is pre-trained for ten full-batch update steps. During the main training stage, the model is optimized for 200 full-batch update steps in total. {The regularization coefficient $\alpha$ and the trade-off coefficient $\beta$ are used to balance the contributions of the regularization term and the spectral clustering term in the objective
function, respectively. They are selected from small candidate sets and then fixed as reference settings in the experiments. In practice, the model performance remains reasonably stable within a moderate range around the adopted settings.}

\subsubsection{Evaluation Metrics}
Six widely-used metrics are employed.
The \textbf{\textit{external metrics}}, i.e., Accuracy (ACC), Normalized Mutual Information (NMI), and Adjusted Rand Index (ARI) \cite{metrics}, measure the consistency between clustering results and ground-truth labels. 
The \textbf{\textit{internal metrics}}, i.e., Silhouette Coefficient (SC) \cite{metric/sc}, Davies-Bouldin Index (DBI) \cite{metric/dbi}, and Calinski-Harabasz Index (CHI) \cite{metric/chi}, evaluate clustering quality based on intrinsic data properties without relying on labels. 
The value ranges are defined as: ACC, NMI $\in [0,1]$; ARI, SC $\in [-1, 1]$; and DBI, CHI $\in [0, +\infty)$. 
For all metrics except DBI, a higher value indicates better clustering performance.

\subsection{Quantitative Results}
\subsubsection{Clustering Performance}

\begin{table*}[!t]
\centering
\caption{Clustering performance on the Attributed Graph Datasets evaluated by ACC, NMI, and ARI (Mean$\pm$Std). The best and second-best results on each dataset are marked in \textbf{bold} and \underline{underline}, respectively. `N/A' indicates that the result is unavailable due to gradient explosion. `AR' denotes the average performance rank of each method.}
\resizebox{1\linewidth}{!}{
\begin{tabular}{c|c|rrrrrrrrrrrrrr}
\toprule
    \multirow{2}{*}{Dataset}   
  & \multirow{2}{*}{Metric}  
  & \multicolumn{1}{c}{$K$-Means} 
  & \multicolumn{1}{c}{GAE}
  & \multicolumn{1}{c}{ARVGAE\cite{ARGAE/ARVGAE}}
  & \multicolumn{1}{c}{DAEGC\cite{DAEGC}} 
  & \multicolumn{1}{c}{CCGC\cite{CCGC}} 
  & \multicolumn{1}{c}{DFCN\cite{DFCN}} 
  & \multicolumn{1}{c}{EGAE\cite{EGAE}} 
  & \multicolumn{1}{c}{CONVERT\cite{mm/convert}} 
  & \multicolumn{1}{c}{SCDGN\cite{SCGDN-mm}} 
  & \multicolumn{1}{c}{MAGI\cite{MAGI-kdd}} 
  & \multicolumn{1}{c}{GLAC\cite{GLAC-TAI}}
  & \multicolumn{1}{c}{{DESE\cite{DeSE}}}
  & \multicolumn{1}{c}{{CDC\cite{cdc}}}
  & \multicolumn{1}{c}{\textbf{\model}} \\
  
  &
  & \multicolumn{1}{c}{(Classical)}
  & \multicolumn{1}{c}{(Classical)}
  & \multicolumn{1}{c}{(IJCAI'18)} 
  & \multicolumn{1}{c}{(IJCAI'19)} 
  & \multicolumn{1}{c}{(AAAI'23)} 
  & \multicolumn{1}{c}{(AAAI'21)} 
  & \multicolumn{1}{c}{(TNNLS'22)} 
  & \multicolumn{1}{c}{(MM'23)} 
  & \multicolumn{1}{c}{(MM'23)} 
  & \multicolumn{1}{c}{(KDD'24)} 
  & \multicolumn{1}{c}{(TAI'25)} 
  & \multicolumn{1}{c}{{(KDD'25)}}
  & \multicolumn{1}{c}{{(TNNLS'25)}}
  & \multicolumn{1}{c}{\textbf{(Ours)}} \\

\midrule
\multirow{3}{*}{ACM}
        & ACC &  36.78±0.01  &  44.22±4.11  &  86.94±1.37  &  74.61±10.00  &  89.26±0.60  &  86.04±2.18  &  85.54±3.62  &  80.53±2.91  &  81.64±0.37  &  \underline{89.37±0.51}  &  66.94±12.16  &  {87.42±2.86}  &  {75.19±16.05}  &  \textbf{90.37±0.41}  \\
        & NMI &  0.82±0.01  &  14.67±4.53  &  58.20±3.29  &  47.92±10.35  &  65.36±1.21  &  59.66±4.51  &  56.09±8.26  &  47.45±4.35  &  51.63±0.46  &  \underline{66.00±1.07}  &  34.18±12.69  &  {61.29±4.60}  &  {53.51±15.68}  &  \textbf{67.50±1.17}  \\
        & ARI &  0.24±0.01  &  3.66±2.39  &  64.94±3.29  &  48.70±12.59  &  71.06±1.37  &  63.94±4.79  &  62.10±8.21  &  51.30±6.03  &  54.02±0.37  &  \underline{71.22±0.51}  &  34.03±12.16  &  {66.66±6.35}  &  {54.54±19.97}  &  \textbf{73.57±1.06}  \\
\midrule
\multirow{3}{*}{WIKI}
        & ACC &  25.81±0.89  &  33.11±2.08  &  44.47±3.66  &  25.38±3.35  &  51.29±0.84  &  43.10±3.67  &  47.49±1.13  &  51.41±1.15  &  44.66±0.20  &  \underline{52.61±1.50}  &  23.88±1.66  &  {35.26±0.00}  &  {43.69±4.23}  &  \textbf{52.95±0.88}  \\
        & NMI &  22.69±1.21  &  31.62±1.51  &  44.13±2.65  &  15.15±2.63  &  46.19±1.01  &  38.33±2.91  &  43.33±1.99  &  48.46±0.62  &  44.02±0.55  &  \textbf{51.93±0.68}  &  14.86±2.75  &  {38.39±3.25}  &  {43.37±2.70}  &  \underline{49.26±1.32}  \\
        & ARI &  2.54±0.32  &  5.61±0.89  &  24.44±3.24  &  7.68±2.25  &  25.50±2.72  &  17.17±3.75  &  28.99±1.58  &  28.39±1.33  &  24.33±0.20  &  \textbf{34.61±1.50}  &  6.52±1.66  &  {25.25±2.02}  &  {13.58±3.29}  &  \underline{33.77±0.87}  \\
\midrule
\multirow{3}{*}{CITE}
        & ACC &  26.10±1.33  &  32.93±3.01  &  54.37±2.96  &  42.66±4.74  &  66.31±2.27  &  42.37±2.05  &  58.71±3.68  &  62.14±1.53  &  59.26±0.40  &  66.52±0.70  &  38.47±3.96  &  {39.42±32.18}  &  {\textbf{67.39±4.97}}  &  \underline{66.57±1.14}  \\
        & NMI &  6.92±1.36  &  20.11±2.63  &  27.54±2.85  &  18.79±3.56  &  40.45±2.68  &  23.90±1.83  &  33.15±2.99  &  34.68±1.78  &  34.70±0.16  &  \underline{41.24±0.88}  &  19.24±5.00  &  {37.47±3.57}  &  {\textbf{42.46±3.22}}  &  40.36±1.21  \\
        & ARI &  0.31±1.93  &  4.64±2.01  &  25.11±3.66  &  16.81±4.38  &  39.12±3.36  &  19.19±2.43  &  31.46±4.61  &  34.69±1.88  &  29.58±0.40  &  40.98±0.70  &  16.64±3.96  &  {36.63±5.12}  &  {\textbf{42.96±4.10}}  &  \underline{41.43±1.94}  \\
\midrule
\multirow{3}{*}{DBLP}
        & ACC &  32.74±0.06  &  46.10±1.43  &  54.97±6.88  &  43.36±4.72  &  54.78±1.97  &  38.91±0.04  &  53.64±1.46  &  54.52±2.37  &  45.42±0.02  &  71.57±1.73  &  36.62±3.17  &  {\textbf{73.41±2.38}}  &  {62.54±4.58}  &  \underline{72.46±2.24}  \\
        & NMI &  2.98±0.01  &  19.71±1.83  &  22.61±5.44  &  11.41±3.55  &  23.81±2.53  &  8.11±0.04  &  18.19±1.07  &  22.33±1.93  &  13.63±0.05  &  \textbf{43.40±0.87}  &  6.02±2.90  &  {\underline{41.19±2.39}}  &  {33.50±2.25}  &  39.12±2.27  \\
        & ARI &  15.31±1.87  &  5.78±0.87  &  17.70±5.12  &  10.40±3.70  &  18.64±1.28  &  6.63±0.02  &  15.07±2.02  &  17.81±1.17  &  8.56±0.02  &  \textbf{43.30±1.73}  &  5.45±3.17  &  {\underline{43.17±4.02}}  &  {27.50±3.60}  &  41.24±2.55  \\
\midrule
\multirow{3}{*}{FILM}
        & ACC &  24.21±0.01  &  25.64±0.02  &  24.31±1.32  &  24.61±0.33  &  26.36±0.11  &  25.91±1.64  &  22.79±0.25  &  \underline{27.43±0.23}  &  26.32±0.01  &  23.62±0.18  &  26.47±0.18  &  {15.47±12.64}  &  {\textbf{27.47±0.52}}  &  26.81±0.65  \\
        & NMI &  0.01±0.00  &  0.09±0.01  &  0.22±0.39  &  0.09±0.03  &  0.15±0.01  &  0.28±0.03  &  0.21±0.08  &  0.79±0.07  &  \underline{1.04±0.00}  &  0.26±0.02  &  0.50±0.17  &  {0.26±0.12}  &  {0.93±0.28}  &  \textbf{1.47±0.22}  \\
        & ARI &  0.00±0.01  &  0.13±0.01  &  0.31±0.51  &  0.15±0.10  &  0.24±0.05  &  0.27±0.04  &  0.17±0.08  &  1.34±0.17  &  0.35±0.01  &  0.18±0.18  &  0.84±0.18  &  {0.14±0.27}  &  {\underline{1.47±0.50}}  &  \textbf{1.78±0.26}  \\
\midrule
\multirow{3}{*}{CORA}
        & ACC &  31.14±3.76  &  49.47±5.76  &  66.72±3.04  &  45.30±5.92  &  72.00±1.77  &  45.94±5.80  &  72.11±1.35  &  66.34±1.80  &  \underline{74.30±0.37}  &  67.78±2.89  &  41.11±4.77  &  {23.71±29.15}  &  {55.86±8.32}  &  \textbf{75.82±1.51}  \\
        & NMI &  6.67±5.28  &  40.86±4.81  &  48.96±2.62  &  25.88±4.35  &  55.02±1.91  &  36.46±3.44  &  52.89±1.17  &  46.84±1.68  &  \underline{56.06±0.39}  &  50.56±1.23  &  23.71±4.32  &  {44.91±3.13}  &  {41.37±6.81}  &  \textbf{59.02±1.20}  \\
        & ARI &  7.83±1.69  &  22.49±7.27  &  42.80±2.69  &  20.09±5.99  &  49.17±2.40  &  23.95±5.52  &  48.49±2.16  &  40.13±1.67  &  \underline{51.75±0.37}  &  44.45±2.89  &  18.38±4.77  &  {36.70±3.97}  &  {30.24±9.24}  &  \textbf{55.64±3.06}  \\
\midrule
\multirow{3}{*}{WISC}
        & ACC &  42.03±2.04  &  42.11±1.73  &  37.17±2.58  &  25.38±3.35  &  44.14±0.86  &  40.95±5.44  &  37.01±2.01  &  \textbf{47.61±1.91}  &  45.78±1.03  &  36.97±2.64  &  \underline{47.05±1.39}  &  {46.37±2.55}  &  {39.24±5.19}  &  44.46±1.58  \\
        & NMI &  6.25±1.13  &  8.09±0.63  &  5.35±3.26  &  \underline{15.15±2.63}  &  8.39±0.45  &  7.01±0.58  &  11.02±1.16  &  9.70±3.35  &  11.09±1.90  &  11.52±1.18  &  5.23±1.15  &  {11.15±3.71}  &  {4.39±0.89}  &  \textbf{16.31±1.66}  \\
        & ARI &  -3.02±1.68  &  2.85±0.54  &  2.02±1.63  &  \underline{7.68±2.25}  &  3.60±0.90  &  4.45±0.99  &  6.00±1.12  &  4.76±2.69  &  5.39±1.03  &  6.19±2.64  &  1.94±1.39  &  {6.11±2.88}  &  {-0.88±2.07}  &  \textbf{9.92±1.38}  \\
\midrule
\multirow{3}{*}{UAT}
        & ACC &  32.69±0.12  &  44.55±0.07  &  41.85±1.63  &  52.49±1.25  &  47.88±2.69  &  39.33±4.72  &  53.10±0.79  &  \textbf{55.18±1.34}  &  44.99±2.01  &  43.71±0.33  &  46.59±2.59  &  {43.53±0.00}  &  {44.76±3.29}  &  \underline{53.84±0.27}  \\
        & NMI &  20.63±0.63  &  18.61±9.44  &  15.86±2.38  &  21.42±1.29  &  20.63±2.64  &  13.95±1.67  &  21.80±0.85  &  \textbf{27.31±1.18}  &  12.64±1.32  &  19.18±0.49  &  17.04±2.13  &  {15.36±1.74}  &  {18.55±5.04}  &  \underline{24.15±1.16}  \\
        & ARI &  6.42±0.44  &  11.61±5.90  &  10.33±2.82  &  \underline{21.07±1.11}  &  12.95±1.80  &  7.28±3.16  &  20.77±0.64  &  19.46±1.90  &  11.27±2.01  &  16.45±0.33  &  13.73±2.59  &  {14.44±1.33}  &  {14.32±5.62}  &  \textbf{22.54±0.91}  \\
\midrule
\multirow{3}{*}{AMAP}
        & ACC &  22.66±0.31  &  60.47±0.87  &  N/A  &  47.45±3.36  &  \underline{77.07±0.38}  &  58.51±3.96  &  76.37±1.32  &  66.28±1.86  &  \textbf{77.52±0.02}  &  68.99±0.82  &  42.13±5.07  &  {66.44±12.42}  &  {72.85±6.18}  &  76.02±0.94  \\
        & NMI &  2.37±0.09  &  58.01±0.56  &  N/A  &  38.83±4.24  &  \underline{67.06±0.72}  &  55.95±1.13  &  65.44±1.61  &  52.57±0.79  &  \textbf{70.26±0.69}  &  59.95±0.59  &  27.49±5.71  &  {53.93±14.62}  &  {64.84±3.54}  &  66.47±1.50  \\
        & ARI &  0.37±0.03  &  33.31±1.02  &  N/A  &  25.03±4.68  &  57.55±0.44  &  41.76±1.58  &  57.51±1.74  &  42.89±1.49  &  \textbf{61.21±0.02}  &  47.88±0.82  &  19.51±5.07  &  {44.78±15.66}  &  {53.24±7.87}  &  \underline{57.73±1.50}  \\
\midrule
- & AR &  {12.7}  &  {10.5}  &  {8.6}  &  {9.7}  &  {4.9}  &  {9.4}  &  {6.4}  &  {5.7}  &  {6.1}  &  {\underline{4.6}}  &  {10.7}  &  {7.0}  &  {6.4}  &  {\textbf{2.0}}  \\
\bottomrule
\end{tabular}
}
\label{tab:atrribute-graph-compared-result}
\end{table*}

\paragraph{Performance on the Mixed-Type Attributed Datasets}
Table~\ref{tab:mixed-data-compared-results} shows that \model achieves the best overall performance with an average rank of 1.1, demonstrating strong and stable clustering effectiveness across diverse mixed-type attributed datasets. More detailed results can be found in our conference version paper~\cite{kdd-QGRL}.

On heterogeneous feature datasets (\ie, MM, HF, BC, AA), 
\model shows clear advantages in most cases. {In particular, on MM, \model shows a clear advantage in ARI, achieving 43.40 versus 34.38 for the best competing method, \ie, DESE.} This result supports the effectiveness of aligned encoding in bridging heterogeneous attribute types and preparing unified representations for downstream learning.
{On pure categorical datasets (\ie, TTT, ZO), \model remains highly competitive and performs especially strongly on TTT with an ARI of 83.81 versus 15.42 for the best competing method, \ie, CDC.} This reflects the benefit of multi-level encoding and graph construction in capturing categorical dependencies. 
On pure numerical datasets (\ie, YE, GI, WI, II), it shows clear superiority through quaternion representation learning, effectively modeling feature relationships for clustering.

\paragraph{Performance on the Attributed Graph Datasets}
Table~\ref{tab:atrribute-graph-compared-result} reports the clustering performance of all compared methods, where the cluster number $k$ is set according to the ground-truth labels.
Overall, \model achieves the best average rank among all methods. It significantly outperforms traditional models and is also competitive with state-of-the-art contrastive learning approaches.

{\model achieves the best results across all metrics on ACM and CORA. In particular, its ARI improves by 2.35 points on ACM and 3.89 points on CORA over the best competing methods. This is mainly because FVP better organizes dispersed attribute information, while QGE strengthens the joint modeling of attribute semantics and graph topology through a shallow propagation scheme. Beyond these two datasets, \model also shows clear advantages over classical baselines (\ie, $K$-Means and GAE) and most graph autoencoder-based methods (\ie, ARVGAE, DAEGC, DFCN, and EGAE), further demonstrating its stronger representation ability. On WIKI, FILM, WISC, and UAT, \model remains highly competitive against recent contrastive learning-based methods (\ie, MAGI, CONVERT, GLAC, CCGC, and SCDGN) and structure-oriented methods (\ie, CDC and DESE). These datasets are relatively noisy or have less reliable graph structures, where explicit discriminative objectives or structural optimization may yield sharper cluster boundaries. Nevertheless, \model still achieves comparable results without explicit contrastive supervision, highlighting the effectiveness of the learned quaternion representations.}

\paragraph{{Statistical Significance Analysis}}
{
To further verify the reliability of the observed improvements, the Wilcoxon signed-rank test is conducted on ACC, NMI, and ARI between \model and the compared methods. The results show that \model achieves statistically significant improvements over all baselines on the mixed-type attributed datasets and remains significant against most baselines on attributed graph datasets. Detailed results are provided in online \href{https://github.com/XinxiiChen/AGREE/blob/main/TETCI_AGREE_Online_Appendix.pdf}{\textcolor{black}{Appendix} I-A}.
}

\begin{figure}[!t]
    \centering \includegraphics[width=1\linewidth]{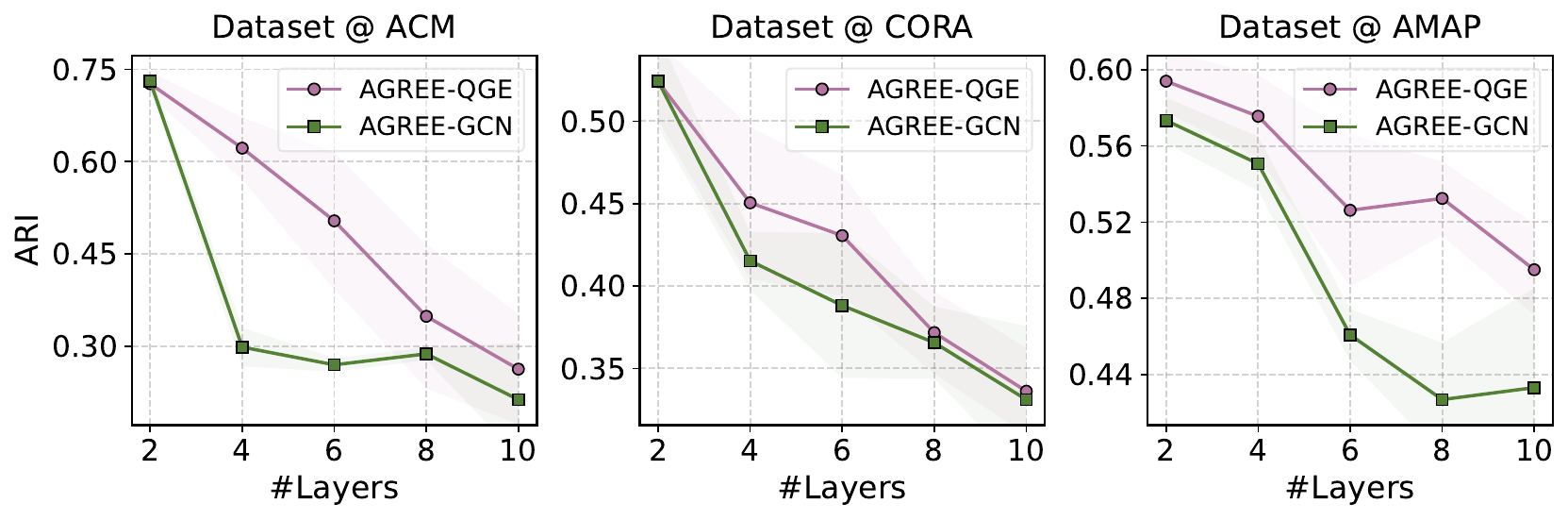}
    \caption{
    {Comparison of ARI between AGREE and the real-valued GCN counterpart on ACM, CORA, and AMAP under increasing encoder depth.}}
    \label{fig:over-smoothing-ablation}
\end{figure}

{To analyze the OS effect, AGREE-QGE is compared with a real-valued counterpart, AGREE-GCN, in which the quaternion graph encoder is replaced by a standard GCN encoder while the other components remain unchanged. Both models are evaluated on ACM, CORA, and AMAP with 2, 4, 6, 8, and 10 graph-encoding layers. Since OS is typically reflected by representation degradation under repeated graph propagation, this comparison allows the effect of increasing depth to be examined more directly. Fig.~\ref{fig:over-smoothing-ablation} reports the ARI results. As the encoder depth increases, both models show clear performance degradation, which is consistent with the OS effect under repeated graph propagation. By contrast, AGREE-QGE degrades more slowly on ACM and AMAP, and remains slightly more stable on CORA. These results suggest that QGE alleviates OS by improving representation learning at shallow depth and reducing the need for deeper propagation.}

\subsubsection{Effectiveness of QGE on Mitigating Over-smoothing (OS)}

{\begin{figure}[!t]
    \centering
\includegraphics[width=1\linewidth]{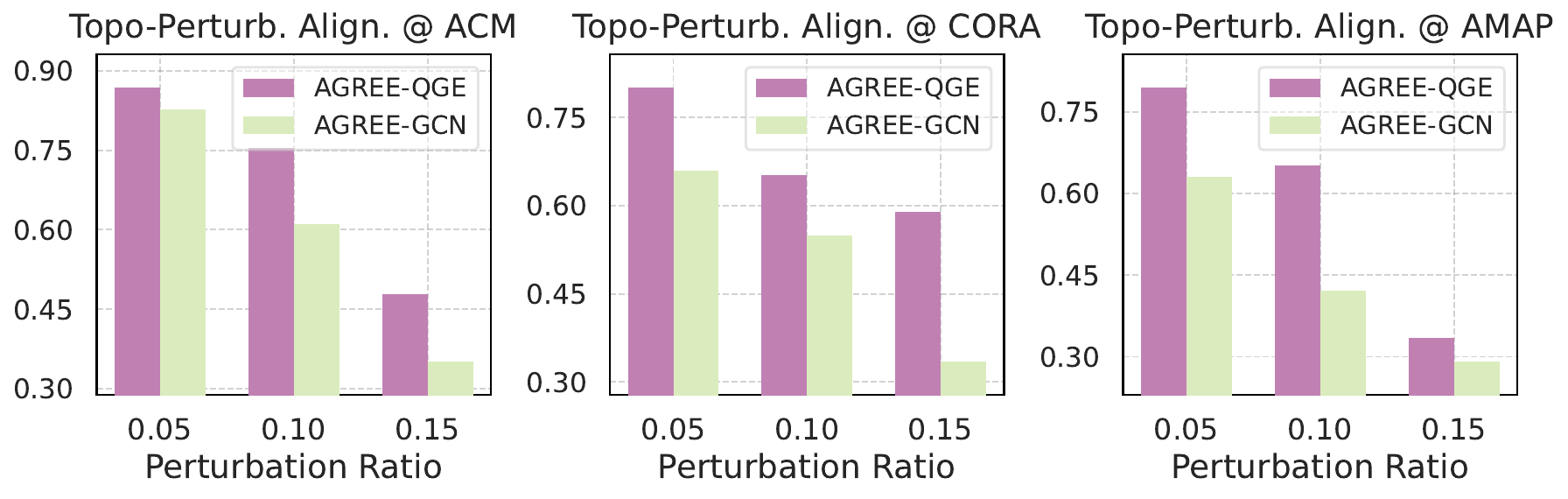}
    \caption{{Comparison of the alignment between the embeddings learned from the original and perturbed graphs for AGREE and its real-valued GCN counterpart on datasets ACM, CORA, and AMAP under different perturbation ratios. Higher values indicate that the learned representation is more robust to topological influence.}}
    \label{fig:od-disturb-cka}
\end{figure}}

\begin{figure}[!t]
    \centering
\includegraphics[width=1\linewidth]{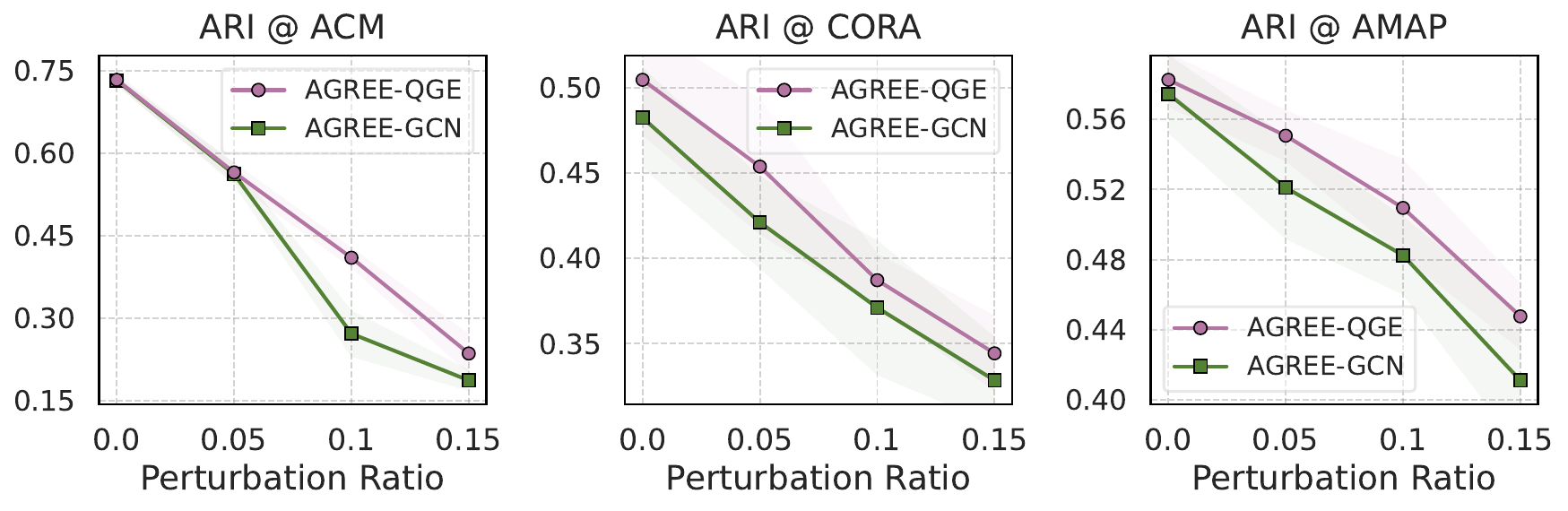}
    \caption{{Comparison of ARI between AGREE and its real-valued GCN counterpart on datasets ACM, CORA, and AMAP under different perturbation ratios.}}
    \label{fig:od-disturb-ari}
\end{figure}

\subsubsection{{Quantitative Analysis of the Over-dominating (OD)}}
{To analyze the OD effect, a topology-perturbation experiment is conducted on ACM, CORA, and AMAP. AGREE and the real-valued counterpart AGREE-GCN are both evaluated with two graph-encoding layers. Cross-class edges are injected into the graph according to each node's degree, with perturbation ratios of 0, 0.05, 0.1, and 0.15. Since OD is reflected by the susceptibility of learned representations to excessive topological influence, incorrect topological perturbations are expected to distort the learned embeddings and clustering results more severely. Accordingly, {topology-perturbation alignment} is measured by CKA~\cite{cka}, which evaluates the similarity between the embeddings learned from the original and perturbed graphs, and {clustering performance} is measured by ARI.}

{Figs.~\ref{fig:od-disturb-cka} and \ref{fig:od-disturb-ari} show that, as the perturbation ratio increases, both representation consistency and clustering performance generally decline on ACM, CORA, and AMAP. This is consistent with the OD effect, since incorrect topology increasingly distorts the learned representations and clustering results. Compared with AGREE-GCN, AGREE-QGE maintains higher representation consistency on most perturbation settings and usually exhibits a slower ARI degradation trend. These results suggest that QGE is less sensitive to incorrect topological influence during representation learning.}

\begin{figure}[!t]
    \raggedright
        \includegraphics[width=1\linewidth]
        {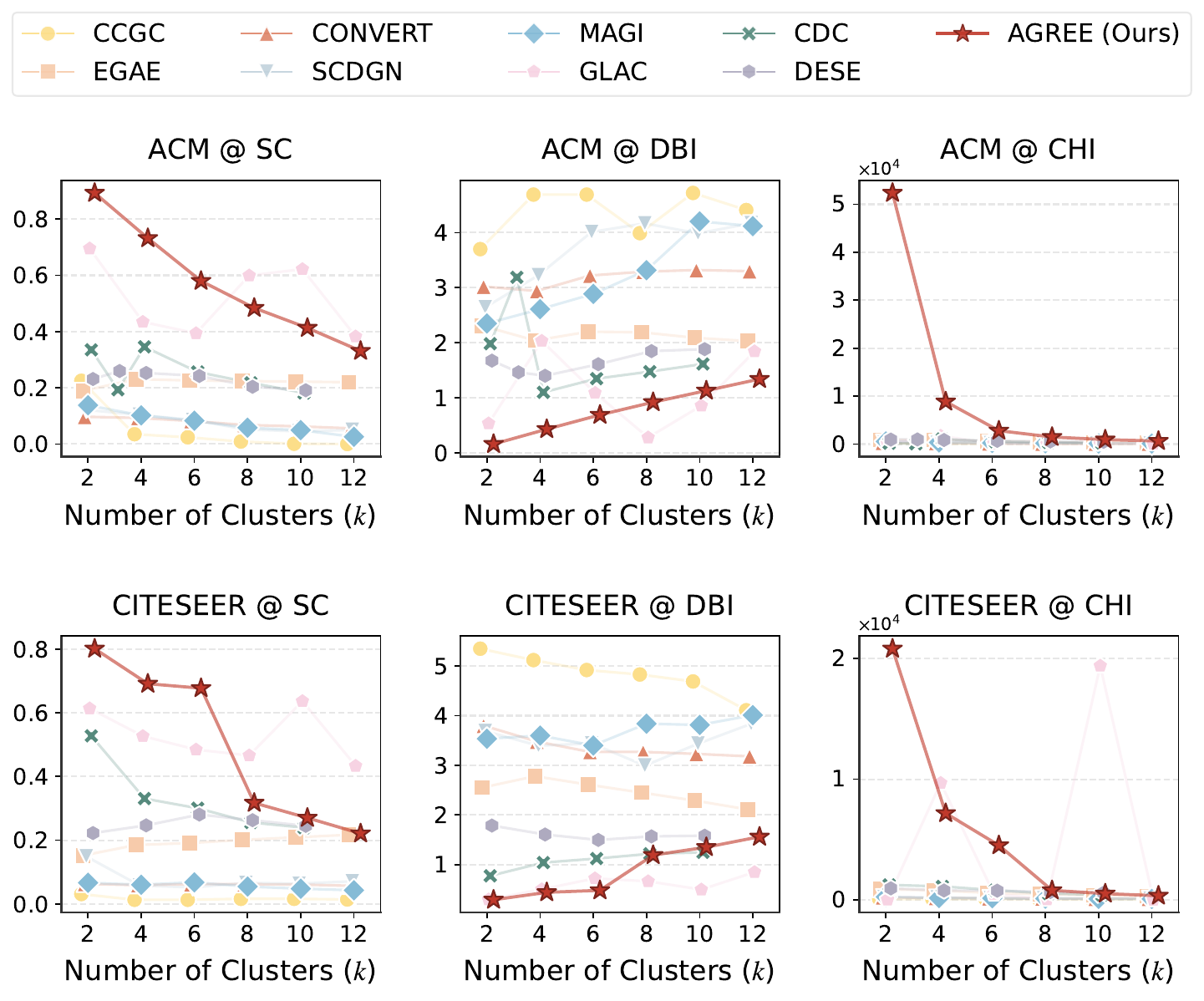}
    \caption{Clustering performance comparison under different numbers of clusters $k$, evaluated by the internal metrics SC, DBI, and CHI. Higher values indicate better performance
  for SC and CHI, while lower values are preferred for DBI.}
    \label{fig:different-k-clustering}
\end{figure}

\subsubsection{Separability and Universality Evaluation of Representations}
 
To assess the effectiveness of \model under varying numbers of clusters, {its performance is compared with the strongest baselines in Table~\ref{tab:atrribute-graph-compared-result} (\ie, EGAE, CCGC, CONVERT, SCDGN, MAGI, GLAC, DESE, and CDC) using internal metrics (\ie, SC, DBI, and CHI).} {As shown in Fig.~\ref{fig:different-k-clustering}, \model shows stronger overall performance over most $k$ ranges and metrics, highlighting the strong separability and robustness of the learned embeddings.} Moreover, since \model does not rely on the ground-truth number of clusters during training, the same learned representation can be directly evaluated under different $k$ values without retraining, which further demonstrates its universality.
More specific observations are as follows: 
1) As $k$ increases, the performance of all methods tends to decline slightly. This is because internal metrics emphasize inter-cluster separation and intra-cluster compactness, while larger $k$ usually produces smaller and more compact clusters, making the metric values gradually converge and become less sensitive to performance differences. 
2) The $k$ values in Figure~\ref{fig:different-k-clustering} include the ground-truth cluster numbers listed in Table~\ref{tab:all-datasets}. At these true $k$ values, \model still maintains superior performance, further demonstrating the strong separability of its learned representations.

\subsubsection{Ablation Studies}

\begin{table}[!t]
\centering
\caption{Ablation study of the key modules of \model on attributed graph datasets. The best and second-best results are marked in \textbf{boldface} and \underline{underline}, respectively.}
\resizebox{\linewidth}{!}{
\begin{tabular}{c|c|ccccc}
\toprule
Dataset & Metric & Baseline & w/o FVP & {w/ Complex GE} & w/o $\beta$ & \textbf{\model} \\
\cmidrule(lr){1-7}
\multirow{3}{*}{ACM}
        & ACC & 89.34 & 89.90 & {88.79} & \underline{90.44} & \textbf{90.53} \\
        & NMI & 64.54 & 65.80 & {63.81} & \underline{67.48} & \textbf{67.67} \\
        & ARI & 70.92 & \underline{72.27} & {69.67} & \underline{73.70} & \textbf{73.93} \\
\cmidrule(lr){1-7}
\multirow{3}{*}{WIKI}
        & ACC & 51.60 & 51.57 & {{52.27}} & \textbf{53.16} & \underline{52.59} \\
        & NMI & {49.15} & 47.91 & {48.68} & \textbf{50.13} & \underline{50.01} \\
        & ARI & {32.43} & 31.99 & {28.31} & \underline{34.24} & \textbf{34.35} \\
\cmidrule(lr){1-7}
\multirow{3}{*}{CITE}
        & ACC & 65.73 & 66.21 & {66.64} & \underline{67.31} & \textbf{67.41} \\
        & NMI & 40.24 & 40.44 & {41.06} & \underline{41.08} & \textbf{41.14} \\
        & ARI & 40.72 & 41.28 & {41.44} & \underline{42.07} & \textbf{42.17} \\
\cmidrule(lr){1-7}
\multirow{3}{*}{DBLP}
        & ACC & 67.89 & \underline{71.41} & {64.70} & 69.93 & \textbf{72.45} \\
        & NMI & 35.20 & \underline{38.15} & {34.60} & 37.80 & \textbf{39.58} \\
        & ARI & 35.48 & \underline{39.73} & {27.90} & 38.15 & \textbf{40.96} \\
\cmidrule(lr){1-7}
\multirow{3}{*}{FILM}
        & ACC & 26.95 & \underline{27.41} & {\textbf{29.51}} & 27.42 & 26.84 \\
        & NMI & 1.12  & 1.28 & {\textbf{2.04}} & \underline{1.49} & 1.39 \\
        & ARI & 1.76  & \underline{1.97} & {\textbf{2.95}} & 1.86 & 1.74 \\
\cmidrule(lr){1-7}
\multirow{3}{*}{CORA}
        & ACC & {72.73} & \underline{73.12} & {71.53} & 70.89 & \textbf{73.28} \\
        & NMI & \underline{55.84} & 55.13 & {55.31} & 53.25 & \textbf{56.22} \\
        & ARI & \textbf{51.44} & {50.61} & {48.43} & 48.05 & \underline{51.34} \\
\cmidrule(lr){1-7}
\multirow{3}{*}{WISC}
        & ACC & 40.92 & 43.03 & {42.47} & \underline{43.63} & \textbf{44.14} \\
        & NMI & 13.54 & 16.09 & {9.86}  & \underline{16.40} & \textbf{16.59} \\
        & ARI & 8.03  & 9.37  & {0.71}  & \underline{9.78}  & \textbf{10.38} \\
\cmidrule(lr){1-7}
\multirow{3}{*}{UAT}
        & ACC & 53.68 & \underline{53.71} & {52.66} & 53.47 & \textbf{53.82} \\
        & NMI & {23.69} & \underline{23.77} & {23.68} & 23.68 & \textbf{24.03} \\
        & ARI & 21.87 & \underline{22.36} & {21.39} & 22.11 & \textbf{22.38} \\
\cmidrule(lr){1-7}
\multirow{3}{*}{AMAP}
        & ACC & 73.23 & 74.69 & {71.42} & \textbf{77.00} & \textbf{77.00} \\
        & NMI & 61.13 & 63.65 & {62.33} & \textbf{69.01} & \textbf{69.01} \\
        & ARI & 53.81 & 55.53 & {51.61} & \textbf{59.85} & \textbf{59.85} \\
\cmidrule(lr){1-7}
- & AR & 4.60 & 3.75 & {4.22} & \underline{2.87} & \textbf{1.88} \\
\bottomrule
\end{tabular}
}
\label{tab:ablation-study-attributed-graph}
\end{table}

\paragraph{Ablation Study on Attributed Graph Data}

Table~\ref{tab:ablation-study-attributed-graph} reports the clustering performance on attributed graph data. Here, Baseline refers to the real-valued counterpart of \model,
consisting of one MLP followed by two GCN encoders. Three additional variants are considered: \model w/o FVP (FVP replaced by a linear layer), {\model w/ Complex GE (QGE replaced
by a complex-valued graph encoder)}, and \model w/o $\beta$ ($\mathcal{L}_{sc}$ removed), together with the full \model model.
Comparisons with these variants reveal the respective roles of FVP, QGE, and the clustering-oriented loss in enhancing attribute preservation, extraction, and adaptability. 
Several observations can be drawn:
1) Adaptability: \model surpasses the Baseline in 26/27 cases, highlighting the adaptability of FVP and QGE in representation learning.
2) Role of FVP: \model outperforms w/o FVP in 26/27 cases, confirming FVP as a necessary pre-phase for QGE. Without FVP, the four MLPs become unlearnable, reducing FVP to a random projection and failing to generate suitable feature quaternions for QGE.
3) Role of QGE: \model exceeds w/o QGE in 24/27 cases.
{It should be noted that this variant is already a stronger substitution than a standard real-valued GCN, since complex-valued representation learning still preserves hyper-complex feature coupling to some extent. Even under this stronger setting, \model remains stronger overall. This suggests that quaternion representation learning provides stronger expressive power than its lower-dimensional hyper-complex counterpart for modeling attribute interactions and alleviating the OD effect.}
4) Cluster-oriented loss: \model outperforms w/o $\beta$ in 24/27 cases, demonstrating that $\mathcal{L}_{sc}$ enhances representation learning by guiding optimization toward more clustering-friendly embeddings.

\begin{table}[!t]
\centering
\caption{Ablation study of the key modules of \model on mixed-type attributed datasets. AE represents the mixed-typed attributed graph aligned encodings, and QRL represents the quaternion representation learning. The best and second-best results are marked in \textbf{boldface} and \underline{underline}, respectively.}
\resizebox{0.9\linewidth}{!}{
\begin{tabular}{c|c|cccc}
\toprule
Dataset & Metric & Baseline & Baseline + AE & Baseline + QRL & \textbf{\model} \\
\cmidrule(lr){1-6}
\multirow{3}{*}{MM}
        & ACC & 54.99 & 53.46 & \underline{68.19} & \textbf{82.96} \\ 
        & NMI & 0.77  & 0.73  & \underline{9.97}  & \textbf{34.87} \\ 
        & ARI & 0.89  & 0.52  & \underline{13.13} & \textbf{43.40} \\ 
\cmidrule(lr){1-6}
\multirow{3}{*}{HF}
       & ACC & 62.61 & \underline{70.80} & 65.28 & \textbf{71.37} \\ 
       & NMI & 2.18  & \underline{6.06}  & 4.04  & \textbf{10.92} \\ 
       & ARI & 5.13  & \underline{13.10} & 8.11  & \textbf{16.63} \\
\cmidrule(lr){1-6}
\multirow{3}{*}{BC}
        & ACC & \underline{70.28} & 52.45 & 54.62 & \textbf{72.03} \\ 
        & NMI & \underline{6.27}  & 1.81  & 0.85  & \textbf{8.66}  \\ 
        & ARI & \underline{13.66} & -0.77 & 0.60  & \textbf{17.00} \\ 
\cmidrule(lr){1-6}
\multirow{3}{*}{AA}
        & ACC & 52.02 & 53.47 & \underline{63.46} & \textbf{63.56} \\ 
        & NMI & 0.13  & 0.73  & \underline{4.49}  & \textbf{4.71}  \\ 
        & ARI & -1.23 & 0.52  & \textbf{3.86}  & \underline{3.75}  \\   
\cmidrule(lr){1-6}
\multirow{3}{*}{TTT}
        & ACC & 60.34 & \underline{61.38} & 61.53 & \textbf{95.74} \\ 
        & NMI & 1.24  & 1.90  & \underline{4.61}  & \textbf{76.33} \\ 
        & ARI & 3.40  & 4.44  & \underline{5.25}  & \textbf{83.81} \\ 
\cmidrule(lr){1-6}
\multirow{3}{*}{ZO}
        & ACC & 65.05 & 53.57 & \underline{70.30} & \textbf{80.20} \\ 
        & NMI & 63.45 & 0.75  & \underline{75.71} & \textbf{82.12} \\ 
        & ARI & 49.88 & 0.53  & \underline{56.17} & \textbf{77.33} \\ 
\cmidrule(lr){1-6}
-      & AR  & 3.28  & 3.28  & \underline{2.33}  & \textbf{1.11}  \\
\bottomrule
\end{tabular}
}
\label{tab:ablation-mixed-typed-attributed}
\end{table}

\paragraph{Ablation Study on Mixed-Type Attributed Data}

\begin{figure}[!t]
  \centering
  \includegraphics[width=0.9\linewidth]{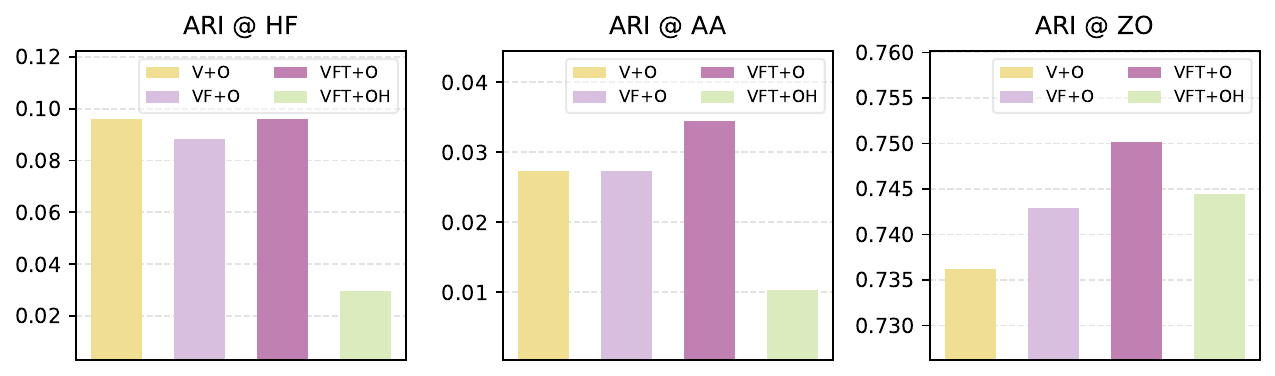}
  \caption{{Comparison of ARI for the progressive alignment ablation on mixed-type attributed data. Here, \textit{V}, \textit{F}, and \textit{T} denote value-, feature-, and attribute-type encoding, respectively. \textit{O} denotes the proposed object-level graph construction, while \textit{OH} denotes the one-hot similarity graph.}}
  \label{fig:alignment-ablation-ari-selected}
\end{figure}

Table~\ref{tab:ablation-mixed-typed-attributed} reports the ablation results on mixed-type attributed data. Here, Baseline refers to the GAE, equipped with KL loss, one-hot encoding, and Euclidean distance-based graph construction (Remark~\ref{remark:Euclidean distance}). Three additional variants are considered: Baseline+AE (with the aligned encoding module), Baseline+QRL (with the quaternion representation learning module), and the full \model model.
Several observations can be drawn:
1) Baseline+AE yields no notable improvement, as AE mainly captures hierarchical couplings, which alone are insufficient to enhance representation learning and may introduce redundancy without a complementary mechanism.
2) Baseline+QRL achieves superior performance on datasets such as AA and MM, reflecting QRL’s ability to extract meaningful features from data with limited information richness. However, its performance declines on BC, likely due to the absence of AE, which limits hierarchical coupling encoding and hinders QRL’s pattern learning.
3) The advantage of \model over Baseline+QRL is marginal on AA because most categorical features are Boolean-valued and largely independent, diminishing AE’s effect. Since both models use QRL, which improves representation flexibility, their performance remains similar on AA while exceeding that of Baseline and Baseline+AE.
(4) Overall, the full \model achieves substantial gains across most datasets, demonstrating that AE and QRL are mutually complementary: AE provides structured encoding, while QRL enhances coupling learning, together producing more comprehensive and discriminative representations for clustering.

{To evaluate the contribution of each alignment level, an ablation study is conducted. Specifically, the aligned encoding is introduced incrementally from \textit{V}alue-level encoding to \textit{F}eature-level and attribute-\textit{T}ype encodings on the proposed \textit{O}bject-level graph. For comparison, the proposed graph \textit{O} is also replaced with a \textit{O}ne-\textit{H}ot similarity graph to assess its contribution.
{Fig.~\ref{fig:alignment-ablation-ari-selected} reports the
corresponding clustering performance in terms of ARI. 
On AA and ZO, \textit{VFT+O} achieves best performance, highlighting the value of the richer alignment design in AGREE. On HF, performance remains similar across different encoding variants, suggesting that lower-level statistics are already sufficient. However, replacing the proposed object-level graph \textit{O} leads to clear degradation on all three datasets, confirming the effectiveness of the proposed graph construction in alleviating attribute heterogeneity and preserving clustering-relevant structure. Overall, although the benefit of richer alignment varies across datasets, the full aligned encoding provides a more complete and generally effective design for mixed-type attributed data clustering.}}

\subsubsection{{Efficiency Analysis}}  

To evaluate the efficiency of \model under increasing graph size, node-scaling experiments are conducted on sampled subgraphs of ogbn-arxiv from the Open Graph Benchmark (OGB)~\cite{hu2020ogb}. The results show that \model scales steadily with graph size and maintains competitive runtime and memory efficiency, demonstrating acceptable scalability under the evaluated settings. Detailed experimental settings and results are provided in online \href{https://github.com/XinxiiChen/AGREE/blob/main/TETCI_AGREE_Online_Appendix.pdf}{\textcolor{black}{Appendix} I-B}.

\subsection{Qualitative Results}
\begin{figure}[!t]
    \centering
\includegraphics[width=1\linewidth]{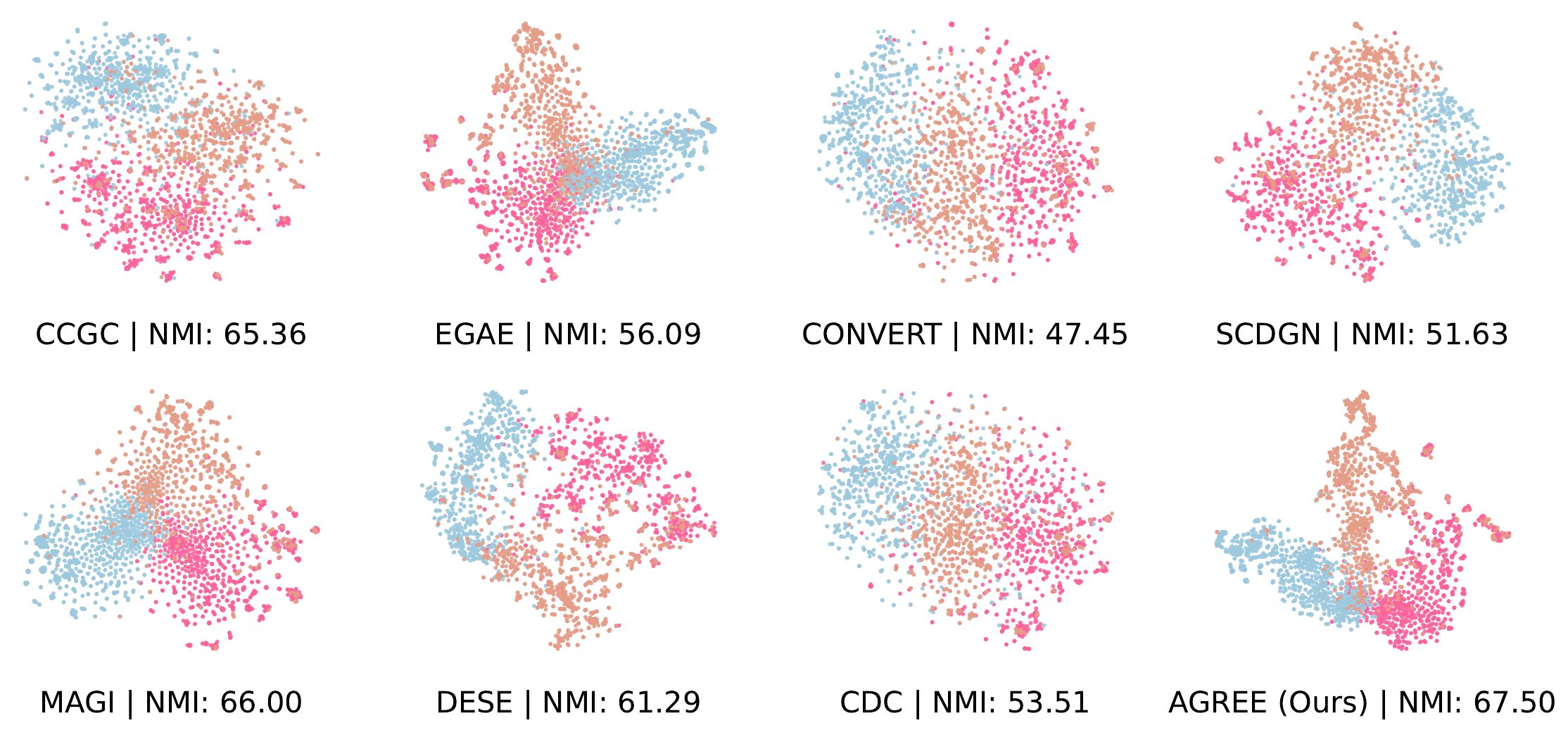}
\caption{$t$-SNE visualization of the ACM dataset produced by eight representative methods and \model. Different colors denote ground-truth clusters, and the corresponding NMI values are also reported.}
\label{fig:acm-graph-dataset-visual}
\end{figure}

To provide an intuitive view of the learned representations, {Figure \ref{fig:acm-graph-dataset-visual} visualizes the embeddings generated by several competitive methods on ACM (\ie, CCGC, EGAE, CONVERT, SCDGN, MAGI, DESE, CDC, and \model) using $t$-SNE~\cite{tsne}.} Different colors denote the ground-truth clusters, and the corresponding NMI values are also reported. Overall, \model, MAGI, and CCGC exhibit the clearest cluster separation, which is consistent with their performance on the external metrics. Among them, \model shows the most favorable balance between intra-cluster compactness and inter-cluster separability. {EGAE and DESE also produce relatively compact local groups, but their global cluster boundaries are less distinct.} By contrast, the remaining methods show more overlap among clusters, indicating weaker discriminative structure in the learned embeddings. These observations are consistent with the quantitative results on ACM and further suggest that \model learns more clustering-friendly representations by better integrating attribute semantics with graph structure.

\section{Conclusion}\label{sec:conclusion}
This paper proposes AGREE, a novel any-type attributed graph clustering method for unified representation learning on attributed graphs and heterogeneous attribute data. To reduce attribute heterogeneity and improve the coordination between attributes and graph structure, AGREE introduces a multi-level aligned encoding scheme together with similarity graph construction to improve the comparability and organization of mixed-type attributes for subsequent learning. It then incorporates quaternion representation learning to strengthen structured feature interaction. Specifically, Four-View Projection organizes arbitrary-dimensional attributes into quaternion-compatible inputs, and the subsequent quaternion graph learning process enhances cross-view interaction and attribute-side modeling, thereby helping alleviate the OD effect. Its shallow yet expressive architecture also helps alleviate the OS effect. Under the joint optimization of graph reconstruction and clustering objectives, AGREE learns embeddings with improved semantic consistency and clustering discrimination. This work also identifies the OD effect in graph clustering, offering a useful perspective for understanding topology-dominated representation behavior. In addition, AGREE extends quaternion-based graph learning to more general attributed data settings, and its learned embeddings can be reused under different candidate $k$ values without retraining. Experimental results on diverse benchmarks demonstrate the effectiveness, robustness, and adaptability of AGREE. 
Future work will extend to more complex graph data forms and more challenging real-world settings, such as federated graph learning. Potential bias introduced by skewed or imbalanced attributes in heterogeneous clustering will also be explored to broaden practical impact.

\bibliographystyle{ieeetr}
\bibliography{ref}

\end{document}